\documentclass[conference]{IEEEtran}
\IEEEoverridecommandlockouts
\usepackage{cite}
\usepackage{amsmath,amssymb,amsfonts}
\usepackage{algorithmic}
\usepackage{graphicx}
\usepackage{textcomp}
\usepackage{xcolor}
\usepackage{booktabs}
\usepackage{xcolor}
\usepackage{rotating}
\usepackage{comment}
\usepackage{multirow}
\usepackage{array}
\usepackage{subcaption}
\usepackage{algorithm}

\def\BibTeX{{\rm B\kern-.05em{\sc i\kern-.025em b}\kern-.08em
    T\kern-.1667em\lower.7ex\hbox{E}\kern-.125emX}}
\begin{document}

% \title{Breaking the Memory Wall for Heterogeneous Federated Learning via Stitching Pre-Trained blocks\\
\title{Heterogeneity-Aware Coordination for Federated Learning via Stitching Pre-trained blocks }

% {\footnotesize \textsuperscript{*}Note: Sub-titles are not captured in Xplore and
% should not be used}

\author{
\IEEEauthorblockN{ Shichen Zhan\textsuperscript{$\dagger \ddag$}, Yebo Wu\textsuperscript{$\dagger \ddag$}, Chunlin Tian\textsuperscript{$\ddag$}, Yan Zhao\textsuperscript{$\S$}, Li Li\textsuperscript{*$\ddag$}} \\
\IEEEauthorblockA{\textsuperscript{$\ddag$}State Key Laboratory of Internet of Things for Smart City,   University of Macau} \\
\IEEEauthorblockA{\textsuperscript{$\S$} Bytedance Inc.}

\thanks{\textsuperscript{*}Corresponding author. Email: llili@um.edu.mo}\\
\thanks{\textsuperscript{$\dagger$}Equal Contribution.}

}

\maketitle

\begin{abstract}

Federated learning (FL) coordinates multiple devices to collaboratively train a shared model while preserving data privacy. However, large memory footprint and high energy consumption during the training process excludes the low-end devices from contributing to the global model with their own data, which severely deteriorates the model performance in real-world scenarios. 
In this paper, we propose FedStitch, 
a hierarchical coordination framework for heterogeneous federated learning with pre-trained blocks. 
% a novel FL paradigm that composes the global model from the pre-trained blocks (block). 
 Unlike the traditional approaches that train the global model from scratch,  for a new task, FedStitch composes the global model via stitching pre-trained blocks. Specifically, each participating client selects the most suitable block based on their local data from the candidate pool composed of blocks from pre-trained models. The server then aggregates the optimal block for stitching.  
 % performs a data heterogeneity-aware aggregation and selects the optimal block     
 This process iterates until a new stitched network is generated.
 Except for the new training paradigm, FedStitch consists of the following three core components: 1) an RL-weighted aggregator, and 2) a search space optimizer deployed on the server side, and 3) a local energy optimizer deployed on each participating client.  The RL-weighted aggregator helps to select the right block in the non-IID scenario,  while the search space optimizer continuously reduces the size of the candidate block pool during stitching.  Meanwhile, the local energy optimizer is designed to minimize the energy consumption of each client while guaranteeing the overall training progress. 
%  As a result, FedStitch requires only a small amount of data and introduces no overhead related to backpropagation, thus significantly reducing memory consumption.
% For a specific task or dataset, it relies on Centered Kernel Alignment (CKA) as a method for block selection, incurring only negligible additional inference costs.  
% To mitigate the impact of data heterogeneity in FL, we propose a Reinforcement Learning (RL)-based weighted aggregation algorithm with cross-validation. Additionally, to address the issue of oversized block pools during the block selection process, we have devised an on-the-fly search space reduction method. To minimize the energy consumption on the client, given a round deadline, we propose a real-time local energy coordinator to select the optimal system configuration with minimal energy consumption while meeting the deadline.
% Introducing a model pre-trained on publicly available datasets as the global model can, to a certain extent, reduce computational overhead compared to training from scratch. However, for local users with different datasets and tasks, fine-tuning still requires a significant amount of memory usage. This challenge becomes more pronounced in FL scenarios characterized by data heterogeneity.
% Leveraging pre-trained models, we propose FedStitch, a novel neural network creation paradigm in FL for the downstream tasks, the new neural network is generated by stitching different blocks (one or more sequential layers) from several pre-trained models. 
The results demonstrate that compared to existing approaches, FedStitch improves the model accuracy up to 20.93\%.  At the same time, it achieves up to 8.12$\times$ speedup, reduces the memory footprint up to 79.5\%, and achieves 89.41\% energy saving at most during the learning procedure. 
\end{abstract}

\begin{IEEEkeywords}
Federated Learning, resource-efficient, pre-training. 
\end{IEEEkeywords}

\section{Introduction}
Federated learning (FL) enables large amount of clients to collaboratively train a global model while preserving data privacy, which has been widely used to support different kinds of applications \cite{mcmahan2017communication}. 
In order to obtain high-quality analysis, recently developed DNNs are becoming deeper and wider \cite{gpt3}. At the same time, large memory space and high computing power are required within the training process.   
For example,  training a VGG16 model consumes more than 15 GB of memory \cite{melon}. On the other hand, the available RAM for mobile devices only ranges from 4 to 16 GB \cite{howmuch}, which means a large amount of low-end devices are excluded from participating the training process. 
Therefore, the excluded devices cannot make contribution to the global model with their own local data. The performance of the global model is then severely degraded. Meanwhile, on-device training is highly energy-demanding and badly hurts the battery lifetime of mobile devices.  Thus, high resource consumption during the training process seriously impedes the deployment of FL in real-world scenarios.  
% When conducting training on
% mobile devices, federated learning can be costly because the
% whole learning process requires multiple rounds of communication
% between the central server and the devices before the
% model converges. 
% From the perspective of energy consumption,
% on-device training is highly energy demanding and hurts the
% battery lifetime of mobile devices 

\textbf{Limitation of Prior Arts.}
Several approaches have been proposed to deploy FL on resource-constrained devices. The existing methods can be mainly divided into the following two categories.
% To address this problem, popular researches focus on pruning parts of the global model for each client based on their memory budgets and averaging the resulting local models into a global model.
One direction is to prune the channels/widths of CNN-based global models, forming different sizes of sub-models based on the memory budgets of the participating clients \cite{diao2020heterofl,hong2022efficient,horvath2021fjord}.
However, in this way,  each client only sees parts of the global model, severely compromising the model 
architecture. Therefore, the aggregated global model has degraded performance \cite{zhang2023memory}. Another direction is to adjust the depth of a network, in other words, use a layer-level partition. 
Nevertheless, for a sub-network with only initial global model layers, key parts responsible for deep information, like semantic features, are missing. Training it directly on a dataset poses a challenge to learning only corresponding parts of the global model, such as low-level features.
% However, for a  sub-network that only has the initial few layers of the global model, the related parts responsible for learning deep information, such as semantic features, are missing. Therefore, if this sub-network is trained directly on a dataset, it is challenging to let this sub-network only learn the information of the corresponding part in the global model, such as low-level features. 
During aggregation, mismatches in parameters may arise \cite{nokland2019training}, leading to a decline in model performance.
Thus, considering the memory and energy overhead,  a new learning paradigm that can efficiently deploy FL on resource-constrained devices is critical for FL in real-world deployment.
% reduce the memory footprint while guaranteeing the model accuracy 

\textbf{Our Design. } 
In this work, we try to tackle this issue from a new perspective. 
% Considering the limitations of existing works, we present a new perspective for addressing memory constraints.
With the emergence and improvement of an increasing number of open-source datasets, the pre-trained models are prevalent and readily available  \cite{marcel2010torchvision,wolf2019huggingface}. Introducing pre-trained models and fine-tuning them for downstream tasks provide opportunities to tackle the issue of resource limitation from the following perspectives. First,  Fine-tuning produces lower computational, energy, and memory overhead compared to training models from scratch, as a result, more devices can contribute to the learning process with their own local data. Second, leveraging the knowledge acquired on large public datasets can effectively compensate for the insufficient data on devices with resource constraints, further improving performance on local tasks. 
Thus, composing the global model for a specific task with the pre-trained models can be a promising way to surmount the resource limitation in FL.
% in real-world cases.    
Nevertheless, despite the lower cost of fine-tuning on new tasks, 
it remains impractical in highly resource-constrained and disconnected environments for re-training the whole global network, especially for very large models, such as deep computer vision models. If only a portion of the global network is re-trained, the aggregated global model will exhibit particularly poor performance on downstream tasks, while also incurring substantial re-training costs.
Furthermore, faced with an increasingly diverse set of downstream tasks, a single model trained on open-source datasets often lacks sufficient generalization ability for different tasks, further leading to performance degradation.
%making a pre-trained model cannot be directly deployed in resource-limited devices. 
To tackle these challenges, instead of relying on a single pre-trained model, we can harness multiple pre-trained models with distinct network architectures, each offering varying expressive capabilities across different downstream tasks. However, how to leverage these pre-trained models is a new challenge. Simply selecting different models based on tasks fails to exploit the unique expressive capabilities of each model fully, and fine-tuning the models still incurs high training costs.  Here, we introduce a new paradigm to address the above challenges: 
By splitting different pre-trained networks into multiple blocks, selecting part of blocks, and stitching blocks together, we create a stitched network on new task.  Each part of the stitched network comes from a different pre-trained network, allowing us to utilize the divergent advantages of different pre-trained networks. 
% To address this challenge, a new paradigm for creating neural networks is through composition by stitching together  blocks of different pre-trained models instead of retraining \cite{teerapittayanon2023stitchnet}.
% In this way, the training-related overhead can be reduced.

% fine-tune有什么问题  然后引出stitch 然后下面提stitch有什么challenge 单机和FL上 
% 能耗遗迹DVFS得问题和设计在我们的方法那里提 
% challenge拆开  

In this work, we propose FedStitch, a novel paradigm to address the resource limitation in FL. Specifically, we first partition the pre-trained model into blocks (each comprising one or more consecutive layers). Subsequently, on the local dataset, each participant client compares the compatibility between blocks from the block pool through a set of simple forward inferences based on centered kernel alignment scores to select the optimal block.
On the server side, the uploaded selected blocks are aggregated.
 FedStitch continually selects appropriate blocks and stitches them together until a completely new network is generated. The whole process only introduces a little inference overhead, significantly saving the computation and memory consumption related to training. Hence, FedStitch can replace the fine-tuning process on new tasks and is able to be deployed on most highly resource-constrained devices. 
 
 \textbf{Challenges and Techniques.} However, designing such a new learning paradigm faces the following challenges.

\begin{itemize}
    \item  \textit{ Eliminate the impact of non-IID on block selection.}  The data distribution among clients in FL is highly biased.  In this situation,  using a simple aggregation method like FedAvg \cite{mcmahan2017communication} to aggregate block may result in poor performance for the global stitched model. To address this challenge,  we propose a reinforcement learning (RL)-based weighted aggregator on the server to address the data heterogeneity in FL. With the help of the RL algorithm and cross-validation,  the server selects and aggregates the $right$ block in non-IID scenario. 
    \item  \textit{Oversized search space in the block pool.} Although FedStitch can efficiently generate a well-performance network in the downstream task, the huge block pool leads to an enormous block search space, increasing the aggregation time in each round. To reduce the search space during block selection, on the server side,  we deploy a search space optimizer to continuously reduce the size of candidate block pool in each round, further reducing the computation and energy costs.
    \item   \textit{ Suboptimal energy efficiency during local block selection.}  On typical edge devices, the default DVFS governor often sets the highest frequency for local block selection. However, in FL's diverse system landscape, using the highest frequency speeds up inference but delays overall aggregation, waiting for slower updates. Addressing this, we introduce a client-side, feedback-based frequency configuration method: a local energy coordinator. It allows the server to set deadlines for each client per round.    This coordinator predicts energy use across system settings, choosing the best one for each client to balance real-time response and minimal energy consumption.

\end{itemize}

\indent  To the best of our knowledge, we are the first work utilizing pre-trained models without the need for any training in FL. The experiments demonstrate that our approach requires no training overhead related to back-propagation, significantly reducing the memory and energy consumption on the device, and requires much less data compared to the traditional fine-tuning method. This enables the participation of clients with highly resource-constrained, while the generated neural network exhibits comparable performance to the state-of-the-art methods. Thanks to our design tailored for FL, the generated stitched network largely mitigates the impact of non-IID, concurrently accelerating the entire generation process significantly. The main contributions of this paper are summarized as follows:

\begin{itemize}
    \item  We propose FedStitch, an FL framework that employs a new neural network generation method, splitting pre-trained models into blocks and stitching them together for downstream tasks. It avoids any training-related overhead, and each client requires only a few pieces of data.
    \item  We demonstrate the impact of non-IID on the generated neural network and analyze the reason. We design an RL-based weighted aggregation algorithm with cross-validation that significantly reduces the influence of non-IID and accelerates the entire process with an on-the-fly CKA-based search space reduction method.
    \item  For reducing energy consumption, we propose a feedback-based frequency configuration method to meet the real-time requirement while minimal energy cost.   
    \item Extensive experiments evaluate the effectiveness of FedStitch on accuracy improvement, generation speed-up, and cost reduction for energy and memory. 
\end{itemize}

\section{Related Work}
\subsection{FL on Resource-limited Devices}

Recent research on memory limitations in FL partitions the global model into local sub-models based on each client's memory budget. Approaches like HeteroFL \cite{diao2020heterofl} and FjORD \cite{horvath2021fjord} allow variability in model architecture across clients through diverse model widths/channels.
% akin to slimmable neural networks \cite{yu2018slimmable}. 
Others, like InclusiveFL \cite{liu2022no} and DepthFL \cite{kim2022depthfl}, allocate models of different sizes to clients based on their on-device capabilities by adjusting the network's depth, while FEDEPTH \cite{zhang2023memory} decomposes the full model into blocks and trains blocks sequentially to obtain a full global model. 
ProFL \cite{breaking}  divides the model into
blocks, trains the front blocks, and safely freezes them after convergence.
In summary, existing research primarily focuses on splitting the global model among users based on their hardware constraints. Regardless of the partitioning method, each client receives only a partial global model, with only a few high-memory clients getting the full model. Consequently, locally trained models fail to fully capture features and lack expressiveness. The final aggregated model also suffers from parameter mismatch issues \cite{nokland2019training}, resulting in subpar performance.

In this work, our approach not only eliminates the need for training, thereby avoiding a huge amount of associated overhead, but also utilizes a full pre-trained model, whose knowledge acquired on public datasets can be leveraged on specific tasks. This allows the most of users to access the full global model in the generation process, and the final aggregation model also exhibits high performance for new tasks.

\vspace{-0.5em} 
\subsection{Pre-trained Neural Network in Federated Learning}

Pre-training is common in current deep learning to enhance model performance. However, integrating it with FL is nascent, with few studies focusing on it.
References \cite{wufoundationmodel} find pre-training improves FL and narrows the accuracy gap vs. centralized learning, notably with non-IID client data.
Reference \cite{li2023begin} uses pre-trained models for medical image segmentation, mitigating memory and communication overhead with knowledge distillation \cite{distilling}.
FedPCL \cite{tanfedpcl} employs fixed pre-trained neural networks as backbones in FL, sharing updated class-wise prototypes for client-specific representations.

% Pre-training is prevalent in nowadays deep learning to improve the learned model’s performance. However,  incorporating pre-training with FL is still in a very early stage, with only a few works focusing on it.  
% References \cite{zhuangwhen,wufoundationmodel} conduct a series of analysis experiments, fond that pre-training can not only improve FL, but also close its accuracy gap to the counterpart centralized learning, especially in the challenging cases of non-IID clients’ data.
% Reference \cite{li2023begin} leverages pre-trained models in medical image segmentation task. To mitigate the heavy memory and communication overhead of pre-trained models in FL, they use knowledge distillation \cite{distilling}, distilling the knowledge from the large pre-trained model into a smaller model, which can be used for FL initialization.
% FedPCL \cite{tanfedpcl} leverages multiple neural networks as fixed pre-trained backbones to replace the learnable feature extractor. To customize the general representations generated by these backbones for each client, class-wise prototypes with contrastive learning updating  are shared across the clients and the server. 
Although the previous works related to pre-training play a certain role in addressing non-IID problem and reducing the computational overhead in FL, the generation of their global model still requires training, leading to significant memory overhead and energy consumption for clients. In contrast, FedStitch eliminates the need for training entirely, fundamentally addressing this issue.

% This section describes our proposed FedStitch method.
% Section \ref{overview} gives an overview of FedStitch. 
% Section \ref{stitching} introduces the stitching design in FL and identifies the reasons for performance decline in non-IID scenarios.
% Section \ref{rl} details the RL-based weighted aggregator, and an analysis of memory consumption.
% Section \ref{search space} describes the on-the-fly search space optimizer. 
% Section \ref{energy section} introduces the local DVFS energy coordinator for energy saving.
\section{FedStitch: Overall Learning Paradigm} \label{stitching}
Figure 1 represents the overall learning paradigm of FedStitch,  which can be mainly divided into the following main steps. \textcircled{1} In the initialization stage, given a set of models that are pre-trained on public datasets, we first split the models into blocks. We categorize blocks into three types: starting blocks, originating from the initial layers of pre-trained models; intermediate blocks, which may exist multiple times within a single pre-trained model; and terminating blocks, referring to the classifier of the pre-trained model.  
Then, we distribute the pre-trained models and the pool of candidate blocks to all participating clients.   
 % Given the block pool, we divide all blocks into three categories:  starting blocks, intermediate blocks, and terminating blocks. To generate a new neural network for the target dataset $D$, we will begin from the starting blocks.
 \textcircled{2} In each round, the participating clients receive the current stitched network and block pool.
 \textcircled{3} Given a current network $N_s$, each client searches all candidate blocks in the block pool. Let's assume $B_{nl}$ is a candidate block,
 derived from a pre-trained neural network layer $n$ up to layer $l$. We stitch $N_s$ and $B_{nl}$ together to form a candidate stitched neural network. Next, we perform two neural network inference computations, one is that we pass a batch of data of the local target dataset $D$ to the candidate stitched network, resulting in the activation $X$. The other one is passing the same batch data $D$ to the pre-trained network where the candidate block $B_{nl}$ comes from. It stops at layer $l$, resulting in the activation $Y$. We then measure the compatibility to obtain the score of $B_{nl}$.  These operations are repeated for all candidate blocks.  
 \textcircled{4} From these blocks, $K$ blocks with the highest compatibility score are selected and uploaded with their scores.  
 \textcircled{5} At the server side, all received block combinations are integrated through a voting process, and the blocks with the $K$ highest number of votes are determined as the selection for that round.
 \textcircled{6}  Each selected block is stitched with the current network $N_s$, generating $K$ new stitched networks for the next round. The selection continues until a terminating block is picked, the maximum stitched network depth is reached, or all possible paths are explored.
 During the local block selection,  we only need a batch size of data for inference in total.   Compared to training, the amount of required data is significantly reduced.

% In each round, involved clients  receive a current stitched network from server, and perform block selection on their
% local data, choosing the $K$ blocks with the highest CKA scores to upload.  
 In the composing process, the following two key approaches are employed to 1) measure the compatibility of two pre-trained blocks and 2) stitch two blocks.
% In each training round, given a candidate stitched neural network from the server, each client needs to find a suitable block from its block pool to serve as its successive block.
 % The better the compatibility of the selected block, the higher the performance of the stitched neural network.
 % Thus, the following two approaches are employed to 1) measure the compatibility of two pre-trained blocks and 2) stitch two blocks.
 
 \textbf{Measure the compatibility.}
The compatibility of the selected block, the higher the performance of the stitched neural network. We choose Centered Kernel Alignment (CKA) to measure the compatibility between two blocks and to guide the block selection.

Given two representation $K$ and $L$, their CKA score is calculated by:

\begin{equation}
    CKA(K,L) = \frac{\textbf{HSIC}(K,L)}{\sqrt{\textbf{HSIC}(K,K) \textbf{HSIC}(L,L) }}
\end{equation}

\noindent
where $\textbf{HSIC}$ is Hilbert-Schmidt Independence Criterion \cite{gretton2005measuring}. 
For the linear kernels, $\textbf{HSIC}$ is:

\begin{equation}
    \textbf{HSIC}(K,L) = || \text{cov}(X^T X, Y^T Y) ||^2_F
\end{equation}

\noindent
% Specifically, HSIC first computes the kernel matrices for each sample in the representations. Then, it centers the kernel matrices by subtracting the mean along each dimension. Finally, it calculates the inner product of the centered kernel matrices to obtain the CKA score.

% \subsubsection{Stitching Two Blocks}
\label{stitching}
\textbf{Stitch two pre-trained blocks.} 
Once we have selected an appropriate successive block, the next step is to stitch these two blocks together. Since the two blocks may have different input and output dimensions, we use Moore-Penrose pseudoinverse to create a projection tensor $A_{kj}$ which projects an output tensor $X_{ij}$ of the incoming block to an input tensor $Y_{ik}$ of the outgoing block:

\begin{equation}
    A_{kj} = Y_{ik} X^T_{ij} (X_{ij} X^T_{ij})^{-1}
\end{equation}

\noindent 
where $i$ represents the sample dimension, $j$ refers to the output dimension of the incoming block, $k$ is the input dimension of the outgoing block.  We can find $A$ such that $Y_{ik} = A_{kj} X_{ij}$. 

% \subsubsection{Generating Stitched Neural Networks} \label{generating}

\begin{figure}[t]
\centerline{\includegraphics[width=0.9\linewidth]{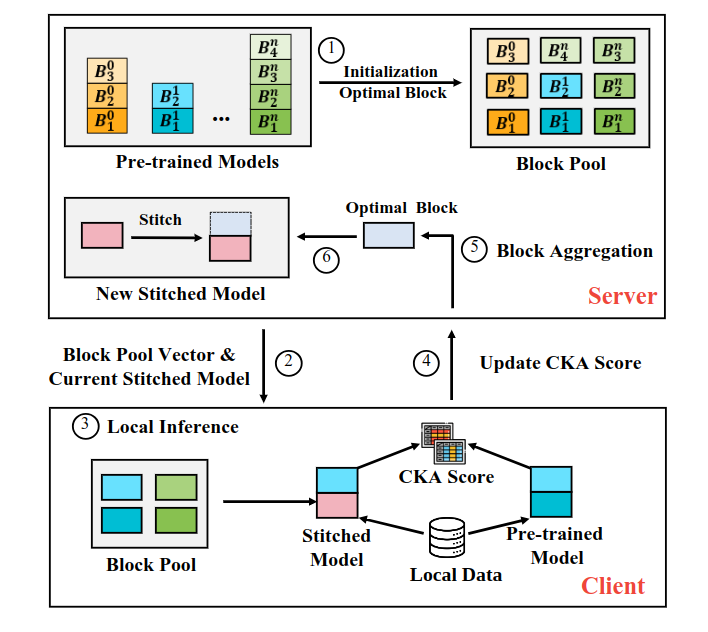}}
\caption{Workflow of Stitched Network Generation.}
\label{fig:generation}
\vspace{-1em}
\end{figure}

However, the data and system heterogeneity in FL pose new challenges. The following key observations motivate the design of the core components of FedStitch.

% \subsubsection{}
% \subsubsection{Underlying Reasons Analysis. 

\textbf{Q1: How does the data heterogeneity impact the stitching performance? }

To investigate the relationship between the distribution of the local training data and the performance of the stitched network, 100 devices with
the same amount of training data and different data distributions are set up. We conduct separate experiments for IID and non-IID scenarios. For the experiment of IID,
each dataset is evenly distributed among all clients, with each
client having data for all classes. For non-IID, we follow \cite{xu2022fedcorr}
to perform the data partition, for each client, the number of
training samples belonging to each class is based on Dirichlet
distribution \cite{frigyik2010introduction} using a concentration parameter set to $\alpha$ = 1. These two groups execute the complete process of generation of the stitched models.
The comparison result is shown in Table. \ref{iid-noniid}. We can observe that non-IID data has a significant impact on the performance of the stitched network, resulting in 5.84\%, 12.77\%, and  9.43\% performance drops in CIFAR10, CIFAR100, and CINIC10 datasets.

\begin{table} [t]
\centering
    \caption{Performance comparison (top-1 accuracy) in IID/non-IID scenarios.}
        \begin{tabular}{c|ccc}
    \toprule
 Experient &  CIFAR10 & CIFAR100 & CINIC10  \\      
    \midrule
 IID &   90.49 &   59.89 & 75.66 \\
 
 Non-IID  & 84.65 & 47.12 & 66.23 \\
    \bottomrule     
  \end{tabular} 
      \label{iid-noniid}
    \vspace{-2em}
\end{table}

\begin{figure}[t]
  \centering
  \begin{subfigure}[t]{0.24\textwidth}
    \centering
    \includegraphics[width=\textwidth]{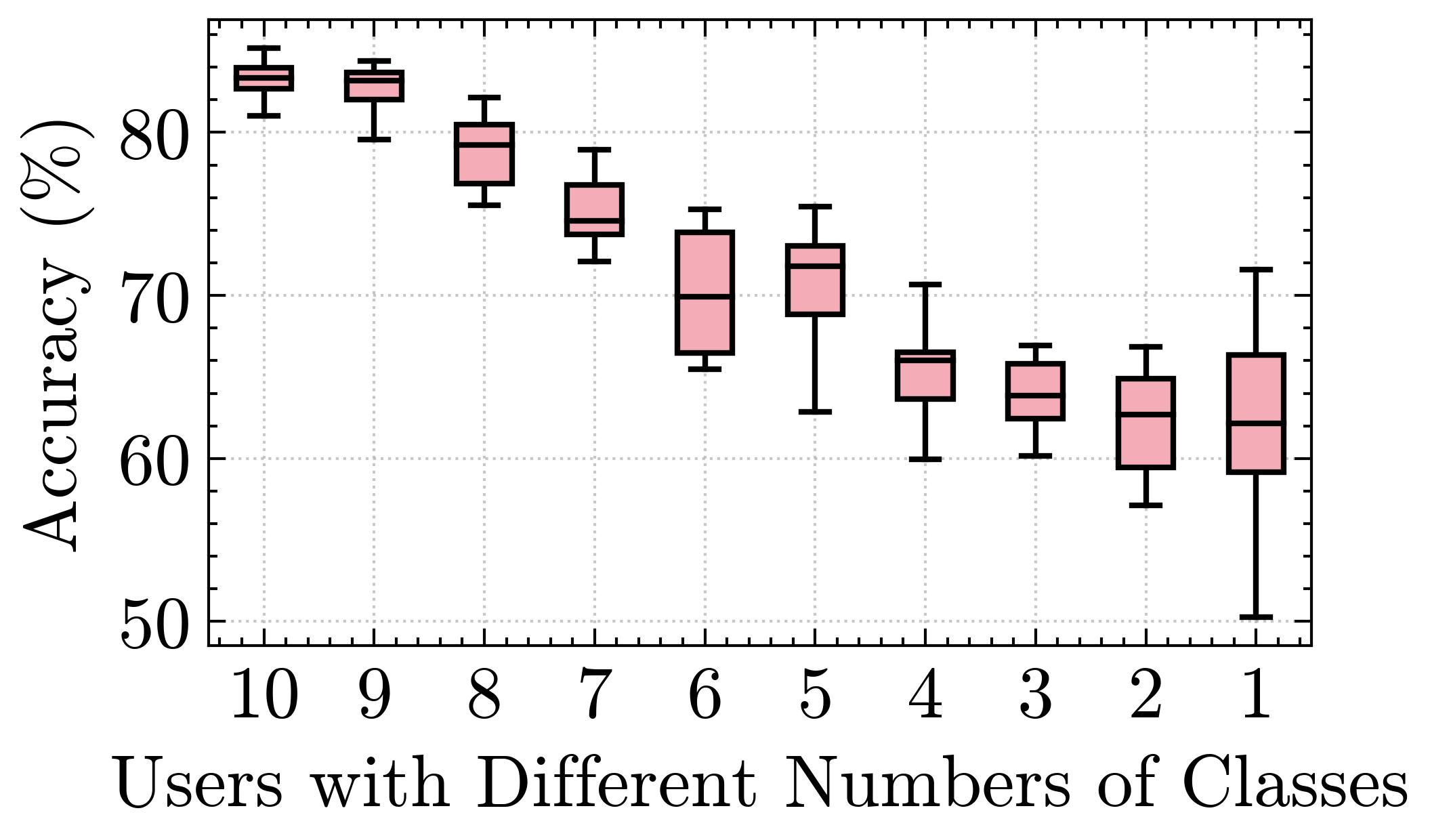}
    \caption{Model performance with different levels of non-IID. }
    \label{fig1}
  \end{subfigure}
  \hfill
  \begin{subfigure}[t]{0.24\textwidth}
    \centering
    \includegraphics[width=\textwidth]{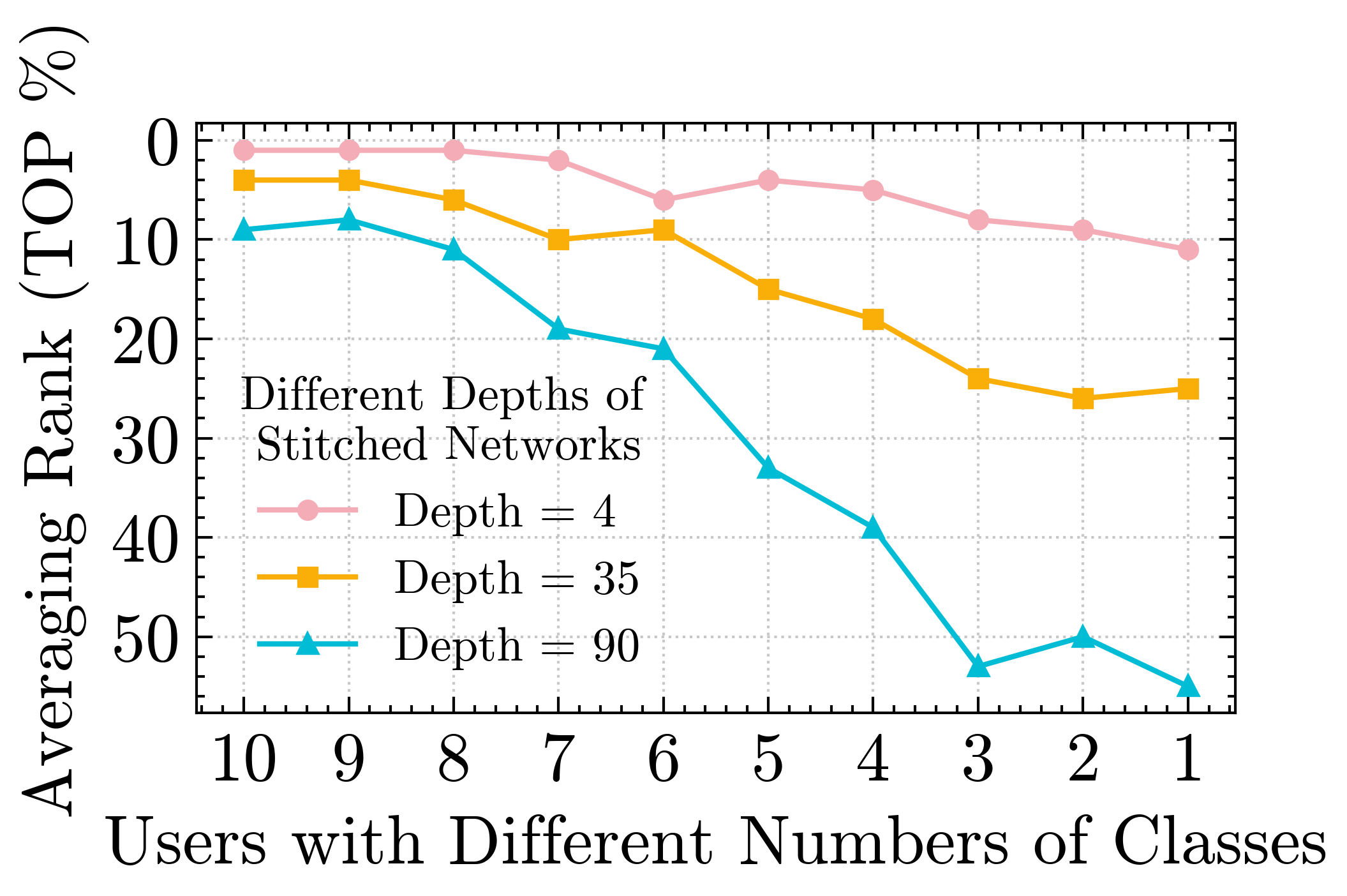}
    \caption{CKA ranking with different levels of non-IID.}
    \label{fig2}
  \end{subfigure}
 \caption{Motivation experiments for statistical heterogeneity (CIFAR10). `Number of Classes' represents the clients with different numbers of classes (`10' means the client has data with all classes). }
 \label{motivation-q1}
 \vspace{-1em}
\end{figure}

To investigate the reasons for degradation, we analyze as follows. 
% Initially, 10 users were established, each with CIFAR10 class data. They independently stitched networks. Best-performing networks from each user were compared (Table \ref{diff class}). Significant performance variations were observed across classes. For example, `dog' outperformed `frog'. Thus, in non-IID scenarios, if most users have dominant class data locally, aggregated network performance on the server will be affected.
% To investigate the reasons behind this degradation, we conduct the following analysis. First,  10 users were set up, each with data from a single class on CIFAR10. They independently completed the stitching process. The best-performing stitched networks from each user were then compared, with results presented in Table \ref{diff class}. we can observe the significant variations in the performance of stitched networks generated from data of different classes. For instance, the performance of stitched networks from the `dog' class is higher than that from `frog'. Therefore, in a non-IID scenario, if the majority of users have dominant data from a specific class in their local datasets, the performance of the stitched network generated through voting aggregation in the server will inevitably be affected.
 We introduced 10 users, each having data of different number of classes, with the same total amount of data. Given the varying impacts of different classes on the results, we conducted 10 sets of experiments with different initializations for each user.
The result is shown in Fig. \ref{fig1}. The results indicate that the lower the non-IID data held by a user (i.e., data encompassing a broader range of classes), the more effectively the produced stitched network performs. Conversely, the performance is still significantly affected for users with only a few classes, even with a sufficient quantity. Therefore, during server aggregation, we should not employ traditional aggregation methods, such as FedAvg \cite{mcmahan2017communication} but rather prioritize users with lower non-IID level data.

% \begin{table}[!ht]
% \centering
%     \caption{Users with different classes in CIFAR10.}
%     % \normalsize
    
%     % \begin{tabular}{c|c|c|c|c|c|c|c|c|c|c}
%         \begin{tabular}{c|ccccc}

%     \toprule
%  Class &  Airplane & Automobile &  Bird   &Cat & Deer   \\   \hline
%   Accuracy (\%) & 59.4 & 60.6 & 59.5   & 65.8 & 59.3  \\   \hline
% Class  & Dog  &  Frog    &  Horse & Ship & Truck\\ \hline
%  Accuracy (\%) & \textbf{69.1} & 57.8 & 61.8 & 62.1   & 62.3\\
%     \bottomrule     
%   \end{tabular} 
%       \label{diff class}

% \end{table}

\textit{\#Principle 1: Local data heterogeneity has a significant impact on the performance of stitched networks, with networks generated from data with lower non-IID levels exhibiting better performance. }

\textbf{Q2: When the server cannot directly access local data, how to select users with lower non-IID level data?}

However, since the server cannot access the local data in FL, it is difficult for the server to identify the client with low non-IID level data. In each round, the server only receives the selected blocks and related CKA scores. Hence, we conduct the following analysis.  Given a candidate stitched network (not yet reaching the terminating block) and selected $N$ users with different levels of non-IID data, we require these users independently to find the most suitable next block from a block pool for this network. We then choose one user $U_k$, record the block he selected, and then look up the CKA score of this corresponding block on another user $U_i$. We calculate the rank of this score within the user $U_i$'s score range of all candidate blocks. We calculate the ranks in the same way for all other users $U_i, i\in \{1,2,...,N\} \setminus \{k\}$. The averaged rank $r_k$ can reflect the performance of $U_k$'s block selection on other users. Except for user $U_k$, we calculate the same rank for each user with the same method as $U_k$. 
To investigate the different impacts of non-IID at different stages of stitching, we configured three candidate stitched networks with varying depths.
In Fig. \ref{fig2}, we observe that low non-IID level users, which have more classes,  will select blocks that always achieve a relatively high CKA rank (not the highest) on other users with different non-IID levels.
% In other words, when a user with data of more classes  achieves a selected block, this block tends to have a relatively high (not the highest) CKA ranking compared to users with fewer classes. 
We will use this observation to identify users with low-level non-IID data in the next section. Additionally, as the stitched network goes deeper, the differences between the blocks selected by users with different levels of non-IID tend to increase. 
% other users  will tend to agree to the block selected by the user with low-level Non-IID data.
% In contrast, users with high-level non-IID exhibit the opposite trend.

\textit{\#Principle 2: Blocks selected by low non-IID level user also have high CKA scores in other users. }

% \textbf{Observation 1. }\textit{ Bias in block selection is observed across varying data classes.}

% \noindent
% \textbf{Observation 2. } \textit{Stitched network generated from IID data exhibits better performance than using non-IID data.}

% \begin{figure}[t]
% \centerline{\includegraphics{Figures/boxplot.png}}
% \caption{Comparison results at different levels of non-IID in CIFAR10. 
% ``Number of Classes" represents the clients with different numbers of classes (`10" means the client has data with all classes). } 
% \label{fig1}
% \end{figure}

% \begin{figure}[t]
% \centerline{\includegraphics{Figures/CKA rank.png}}
% \caption{CKA ranking with different levels of non-IID.}
% \label{fig2}
% \end{figure}

\textbf{Q3: How the system heterogeneity impacts the overall training progress and the stitching overhead of the participating clients? }

To investigate the relationship between the system heterogeneity and the energy consumption of stitching, one user is configured to conduct local block selection at different
fixed process frequencies on the Jetson TX2, Table. \ref{observation-q3} shows the result as an example, we can find that 
 the system spends more than 98\% of the time on
the highest processor frequency during the
local block selection. 
This is because, on edge devices, the default DVFS governor adjusts processor frequency based only on load. When load exceeds a threshold, it picks higher frequencies. Thus, for tasks like DNN on-device inference, it often selects the highest frequency.
% This is because, on a typical edge device, the default Dynamic voltage and frequency scaling (DVFS) governor determines the CPU frequency solely based on the processor load. In scenarios where the CPU load surpasses a predefined threshold, the DVFS governor is designed to opt for higher frequencies. Consequently, for computationally intensive tasks, such as DNN on-device inference, the governor tends to frequently choose the highest available frequency.
However, while this approach ensures the task is completed in the shortest time possible, it may not be energy-optimal for a mobile device in general.  From Table. \ref{observation-q3}, the training time consistently decreases as frequency increases. However, the trend in energy consumption does not align with this pattern. Moreover, due to the system heterogeneity in FL, despite the accelerated inference achieved by employing the highest frequency on
each device, the overall aggregation still needs to wait for the straggler’s updates before turning to the next round. Therefore, for certain highly capable devices, the selection of the highest frequency (as in default DVFS governer) is not imperative.

\begin{table} [t]
\setlength{\tabcolsep}{1.5pt} 
\centering
    \caption{SpeedUp / PowerUp under different DVFS configuration in Jetson TX2.}
\begin{tabular}{cccccc}
    \toprule
 \parbox{1.5cm}{\centering DVFS Configuration} &  \parbox{1.5cm}{ \centering CPU Frequency (GHz)}  & \parbox{1.5cm}{\centering GPU Frequency (GHz)} &   \parbox{1cm}{SpeedUp} &  \parbox{1cm}{PowerUp} & \parbox{1.5cm} {\centering Selection Time (\%)}\\  
    \midrule
 1 &   1.2  & 0.85 & 1$\times$ & 1$\times$ &1.5\\
 2  & 1.4 & 1.12 &  1.21$\times$ & 1.25$\times$ &0.5\\
  3 & 2.0 &1.30 &  1.38$\times$ & 1.14$\times$ &98\\
    \bottomrule     
\end{tabular} 
      \label{observation-q3}
    \vspace{-2em}
\end{table}

\textit{\#Principle 3: The default DVFS governor does not effectively balance the block selection progress and energy consumption in an FL system. }

\section{FedStitch: Core Components}  
Guided by the corresponding principles, in this section,  we present the core components designed to address the challenges introduced by the data and system heterogeneity in FL. 
\subsection{Overview} \label{overview}

\begin{figure}[t]
\centerline{\includegraphics[width=\linewidth]{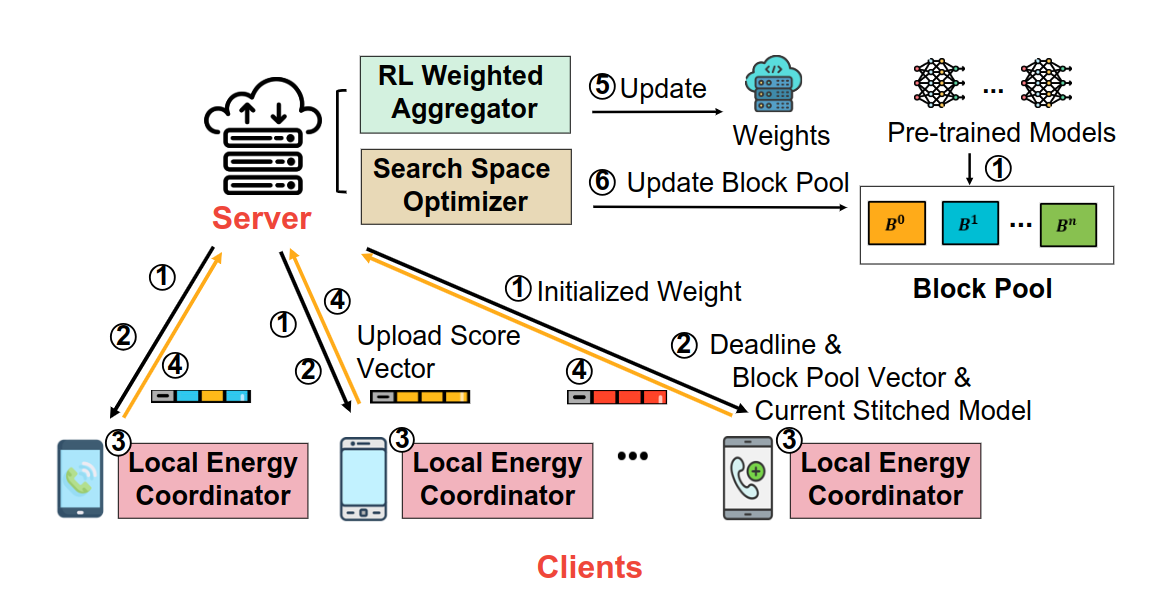}}
\caption{The System Overview of FedStitch.}
\label{fig:overview}
\vspace{-2em}
\end{figure}

Fig. \ref{fig:overview} shows the system overview.  The process is divided into the following steps. 
\textcircled{1} \textbf{Initialization.}  At the beginning, the server assigns initialized weights to each client, and divides the pre-trained models into blocks, forming a block pool. The server then sends both the pre-trained models and the initialized block pool to all participating users.
\textcircled{2} \textbf{Network Dispatching.} In each round, the server dispatches the stitched neural network, block pool state, and deadline to that round's participants.
\textcircled{3} \textbf{Local Selection.}
Locally, the local energy coordinator chooses the optimal configuration for minimizing energy consumption within the deadline. Then, the local user selects the most suitable block for the current stitched neural network using the relevant CKA scores from the block pool.
\textcircled{4} \textbf{Block Updating.} All participants upload their selected block and all associated scores for this round.
\textcircled{5} \textbf{Weighted Aggregation.}
On the server side, the weighted aggregator, utilizing uploaded scores, adjusts each client's weights and employs weighted aggregation to select suitable blocks for the current stitched network. 
\textcircled{6}  \textbf{Space Optimization.}
Simultaneously,  the search space optimizer shrinks the block pool size according to the CKA score results. 
The new stitched network,  weights of clients, and block pool will be updated for the next round as well.
% the RL-based aggregation algorithm with cross-validation to adjust weights for each client. This helps determine the optimal block for the current round, and subsequently, the server updates the weights of clients and block pool with the on-the-fly search space reduction method.
This process repeats until either the terminating block is selected or the stitched neural network reaches its maximum depth.

% \subsection{Generating Stitched Neural Network on Single Device} \label{single}
% \subsection{Generating Stitched Neural Network in FL}  \label{fl}

\subsection{RL Weighted Aggregator} \label{rl}  
% \subsubsection{Reinforcement Learning-based Weighted Aggregation Algorithm with Cross-validation} \label{rl}
Drawing from \textit{Principle 1}, we highlight the significance of users with less non-IID data, suggesting their blocks be prioritized during server aggregation. We propose weighting users' contributions, giving more weight to those with lower non-IID data. However, FL faces hurdles as data privacy prevents server access to this data, complicating weight adjustments. Utilizing \textit{Principle 2}, our approach involves a reinforcement learning-based weighted aggregation to counteract non-IID's effects on block selection. We use the Epsilon-Greedy algorithm to balance performance and exploration in stitching schemes.
% Informed by the key observations  \textbf{Principle 1} in the preceding sections and recognizing the impact of varying levels of non-IID data on the performance of the stitched network, it is advisable to emphasize the importance of users with lower non-IID levels data.
%  This means that blocks proposed by these users should have a higher probability of being selected during aggregation at the server. 
% Naturally, we consider assigning a weight to each user, where a lower non-IID level corresponds to a larger weight. At the server, we adopt a weighted voting approach to select blocks. However, this is confronted with several challenges. In FL, the data cannot leave the local client, preventing the server from accessing data and making it impossible to identify the non-IID level of each user. Additionally, due to incomplete data and potentially unlabeled data on the client, user cannot decide their non-IID level, either. 
% Therefore, adjusting the voting weights becomes difficult.
% Based on the key observation \textbf{Principle 2} that blocks chosen by low non-IID level users tend to have high ranks among other users, we developed a weighted aggregation algorithm using reinforcement learning (RL) and cross-validation to mitigate the impact of non-IID on block selection.
% We exploit the Epsilon-Greedy algorithm from the Multi-Armed Bandit Problem in RL, which ensures the generated stitched network exhibits reliable performance while actively exploring more stitching schemes.  

\begin{algorithm}[tb]
    \caption{ FedStitch }
    \label{alg:fedstitch}
    \textbf{Input}: initialized client weights $W$,  initialized candidate stitched networks set $N$ and related block pool set $B$ \\
    \textbf{Parameter}: exploration factor $\epsilon$, the reward and penalty weight update factors $\alpha$, $\beta$, threshold $\theta$, number of block selection $K$.\\
    \textbf{Output}: updated client weights $W$ and new candidate stitched networks set $N$
    \begin{algorithmic}[1] %[1] enables line numbers
        \FOR{$round \hspace{0.2cm} t=0,1,...,T-1$}
        \STATE $P_t \gets $ Random clients
        \STATE $N_t \gets$  Random candidate stitched network in $N$
        \STATE $B_t \gets $ Related block pool for $N_t$  \\
        % \STATE $S_t \gets \emptyset$ \\
        // Client Score Calculation 
        \FOR{$each \hspace{0.2cm} client \hspace{0.2cm} i \in P_t$}
            \STATE $S_t^i \gets \emptyset$
            \FOR{$each \hspace{0.2cm} block \hspace{0.2cm} b_j \in B_t$} 
                \STATE $s_t^{i,j} \gets $ $ \textit{CKA}(b_j)    $\\
                \STATE $S_t^i \gets S_t^i \cup \{s_t^{i,j}\}$ 
            \ENDFOR
            \STATE $M_t^i \gets$ Select $K$ blocks with highest scores in $S_t^i$ with the probability $1-\epsilon$
            \STATE $M_t^i \gets$ Randomly select $K$ blocks in $S_t^i$ with the probability $\epsilon$
        \ENDFOR \\
        // Server Weighted Aggregation
        \STATE $S_t \gets \bigcup S_t^i \hspace{0.2cm} for \hspace{0.2cm} i \in [1,...P_t] $ 
        \FOR{$each \hspace{0.2cm} client \hspace{0.2cm} i \in P_t$}
            \STATE $r_t^i \gets$ $\textit{RankCalculation}(B_t^i, S_t) $
            \IF{$r_t^i < \theta$}
                \STATE $w_t^i = w_t^i \times (1-\beta) $
            \ELSE
                \STATE $w_t^i = w_t^i \times (1+\alpha)$
            \ENDIF
        \ENDFOR
        \STATE  $W \gets \bigcup w_t^i \hspace{0.2cm} for \hspace{0.2cm} i \in [1,...P_t]$
        \STATE  $M_t \gets \bigcup M_t^i \hspace{0.2cm} for \hspace{0.2cm} i \in [1,...P_t]$
        \STATE $M_t \gets \textit{WeightedVoting}(W,M_t)$
        \STATE $N  \gets N \cup \textit{Stitching}(N_t, M_t) $

        \STATE \textbf{return} $N$, $W$
 
        \ENDFOR
        
    \end{algorithmic}
    
\end{algorithm}

The FedStitch process unfolds as follows:
Initially, every user \(u_i\) is given the same weight \(w_i\), with the sum of their weights equaling 1. We also set the exploration factor \(\epsilon\), and the reward (\(\alpha\)) and penalty (\(\beta\)) update factors, along with a threshold \(\theta\).
In each round, the server sends the current stitched network candidate and block pool to the participating client. The client computes the CKA on their local dataset for each block. Based on the strategy, it either randomly selects \(K\) blocks with probability \(\epsilon\) (exploration) or picks the blocks with the highest CKA scores with probability \(1-\epsilon\) (exploitation).
Clients then upload their selected blocks and all CKA scores. Server-side, cross-validation among users assesses block choices. For a block from \(u_i\), we calculate its average CKA score rank across others. If \(r_i > \theta\), reflecting low non-IID data for \(u_i\), we reward by increasing \(w_i\) with \(w_i \times (1+\alpha)\). If \(r_i < \theta\), indicating high non-IID data, we penalize by reducing \(w_i\) with \(w_i \times (1-\beta)\).  Afterward, the weights are re-normalized to ensure the sum  equal to 1. The server then uses weighted voting to select blocks, stitching them into new candidates. It updates the network set and client weights for the next round.
As stitching progresses, we adjust \(\alpha\), \(\beta\), and \(\epsilon\) to suit different stages. This algorithmic flow is detailed further in our algorithm description.

\addtolength{\topmargin}{0.07in}

\subsection{Search Space Optimizer} \label{search space}

% Although the method proposed in the previous section effectively eliminates the impact of non-IID, the huge block pool leads to an enormous block search space, increasing the aggregation time in each round. For instance, a block pool composed of only five pre-trained neural networks will contain over 50 blocks, and every user has to compare each block in the pool, calculating the CKA score in each round. While the inference cost for CKA is much smaller than training, it still introduces redundant computation, thereby straggling the entire aggregation process. However, we cannot arbitrarily remove blocks from the pool, as we do not know which ones will be valuable for our subsequent selections.
% To address this, we conduct an experiment to demonstrate the ability to dynamically reduce the search space during the stitching process based on CKA scores.
% Considering a pre-trained network divided into six blocks $[B_1, ... B_6]$, if block $B_3$ yields the highest CKA score among the six when selecting blocks for a candidate network, then during the stitching procedure, whether $B_3$ is selected or not, blocks preceding $B_3$ in the pre-trained model ($B_1$ and $B_2$), which contain lower-level features, are less suitable for the new network compared to $B_3$, and thus contribute less to improving the performance of the network.
% Conversely, the blocks containing higher-level features ($B_4$, $B_5$, and $B_6$) are more likely to be selected.
While the proposed method mitigates non-IID effects, the large block pool expands the search space, slowing down aggregation. For example, a pool with five pre-trained networks yields over 50 blocks, necessitating CKA score calculations for each block per round, adding unnecessary computation and delaying aggregation. However, indiscriminate removal of blocks is not feasible without knowing their potential value. To tackle this, we experimented to dynamically narrow the search space using CKA scores during stitching. For a network split into six blocks $[B_1, ... B_6]$, if $B_3$ has the highest CKA score, it implies blocks before $B_3$ ($B_1$ and $B_2$) are less suitable than those after, due to their lower-level features, whereas blocks $B_4$, $B_5$, and $B_6$ are more likely to improve performance due to their higher-level features.

Therefore, when selecting blocks for a candidate stitched network, the client identifies the block with the highest CKA score among those belonging to the same pre-trained model. In the subsequent selections, all blocks of each pre-trained network that are shallower than the block are removed from the block pool.  In every block selection stage, a portion of the block pool is eliminated. The search space progressively reduces during the stitching process, significantly improving 
the speed of each user's block selection
 and accelerating the aggregation process.

\subsection{Local Energy Coordinator}  \label{energy section}

Based on the key observation \textit{Principle 3} that DVFS  governor fails to efficiently balance block selection and energy use in FL, we seek a method to lower energy consumption during local block selection without extending selection time. Simply reducing client frequency during local inference to save energy could delay block selection, especially for slower devices, affecting the entire aggregation process. Therefore, an approach that reduces energy costs without negatively impacting the overall schedule is essential.
% that the default DVFS governor does not effectively balance the block selection progress and energy consumption in an FL system, we need to find a way to minimize the energy consumption during local block selection.  Nonetheless, we cannot merely lower the frequency of each client during local inference to cut down on energy consumption, because it would lead to an increase in the time for local selection. 
% With the computational capabilities among devices differing, a time increase for certain stragglers can influence the entire aggregation's progress. Hence, an approach that can diminish energy cost without detrimentally affecting the overall timeline is required.

To tackle this requirement, we set a deadline for all participating clients in each round. This deadline is determined by the completion time of the last round. With the received hardware configuration information and the size of local input data, the time required by device $i$ to complete the local selection process can be modeled  as: 

\begin{equation}
    t_i =  \frac{c_i D_i}{f_i}  
\end{equation}

\noindent
where $c_i$ represents the number of processor cycles required to
process one data object in inference on mobile device $i$, which can be
obtained through offline profiling, $D_i$ represents the number of
data objects in the local inference data set, and $f_i$ is a particular
process frequency available on device $i$.

With the local completion time of each client $[t_1, t_2, ..., t_N]$ in the last participant round,  the server selects the maximum of them as the deadline for this round $d = max(t_i)$. If the size of candidate stitched networks and related block pools are different from the last round, the deadline $d$ will be adjusted with deep factor $\mu$ and pool size factor $\sigma$.
After receiving the deadline of the current round,  the local energy coordinator conduct a feedback-based system configuration method for each user.  It dynamically adjusts the frequency of the device so that the participant can meet the deadline while minimizing energy consumption.  
To select the proper block for a candidate stitched network,  we model the energy consumption of a device in one
 round  as follows:

\begin{equation}
    E = p^{infer}_{sn} * t^{infer}_{sn} + p^{infer}_{pn} * t^{infer}_{pn} + p^{idle} * t^{idle}
\end{equation}

\noindent where  $ p^{infer}_{sn}$ and $p^{infer}_{pn} $ represent the power consumed while the inference process is running for the current stitched network $sn$ and related pre-trained network $pn$.   $ p^{idle}$ represents the base power when the smartphone is powered on but not actively used.  
$t^{infer}_{sn}$, $ t^{infer}_{pn}$, and $ t^{idle} $  are the time spent in the inference state of two networks and idle state, respectively.

Considering a stipulated deadline $d$ for this round, our goal is to minimize client energy consumption while ensuring the block selection process is finalized before reaching $d$.
Denote the energy consumed by device $i$ during a particular round when running the inference process at a CPU frequency of $f_i$ as $E_i (f_i)$. The problem becomes:

\begin{equation}
    \mathop{\arg\min}\limits_{f_i} E_i (f_i), \quad f_i^{min} < f_i < f_i^{max}
\end{equation}

\begin{equation}
\begin{split}
    \text{s.t.} \quad t^{infe}_{sn,i}(f_i) + t^{infe}_{pn,i}(f_i) +  t^{idle} = d \\ 
    0 \leq t^{infe}_{sn,i}(f_i), t^{infe}_{pn,i}(f_i), t^{idle} \leq d
\end{split}
\end{equation}

% \noindent where $E_i(f_i)$ is the energy consumption  when the
% CPU frequency is $f_i$. 

\noindent
% For each client, we initially allocate a frequency $f_i$ . Subsequently, leveraging the hardware information of the device, we predict its inference time and idle time under this frequency. If the total time is less than the deadline $d$, indicating that $f_i$ is excessively high, we decrement it;  if the total time surpasses  $d$, signifying that $f_i$ is too low, we increment it. According to the feedback of adjustment,  this process continues until the total time stabilizes within a tiny interval not exceeding $d$, at which point we consider the current $f_i$ as the configuration for the client in the current round.
% Using this approach, we dynamically pursue the ideal setup for each client to satisfy the deadline while minimizing energy usage. 
For each client, we assign an initial frequency \(f_i\) and then use the device's hardware specs to estimate its inference and idle times at this frequency. If the combined time falls below the deadline \(d\), suggesting \(f_i\) is too high, we reduce it. Conversely, if it exceeds \(d\), indicating \(f_i\) is too low, we increase it. We adjust \(f_i\) based on this feedback until the total time is just under \(d\), within a narrow margin. This method dynamically fine-tunes \(f_i\) for each client, ensuring we meet deadlines while optimizing energy consumption.

% \begin{table}
% \centering
%   \caption{Statistics summary of generated networks in non-IID on CIFAR10.}  
%   \label{tab:statistics}
    
%   \begin{tabular}{c|ccc}
%     \toprule
%         Model number     & Accuracy (\%) & CKA & Size \\ 
      
%     \midrule
%     FedStitch-3  &  88.17 & 0.87 &  86.2MB \\
    
%     FedStitch-7  &  87.21 & 0.75 &  35.3MB \\
%     FedStitch-1  &  86.95 & 0.89 &  17.8MB \\
%     FedStitch-11  &  84.14 & 0.70 &  101.5MB \\
%     FedStitch-15  &  83.03 & 0.79 &  227.3MB \\
%     \bottomrule     
%   \end{tabular} 
%   \vspace{-1.5em}
% \end{table}

\begin{table*}
\centering
 \large
   % \captionsetup{font=large} 
  \caption{Test results (top-1 accuracy) of various schemes on CIFAR10, CINIC10, and CIFAR100 in IID/non-IID. “FT-Part" means only fine-tuning the classifier of the pre-trained model, almost all clients can participate; “FT-Full" means fine-tuning the whole model, in this case, due to the memory limitation,  the user can participate for 4 pre-trained models ($alexnet$, $resnet50$, $vgg16$, $densenet$, $mobilenet$) are 10\%, 10\%, 10\%, 40\%, 40\% respectively. }  
  \label{tab:result}

  \resizebox{\linewidth}{!}{
   \renewcommand{\arraystretch}{1.5}
  \begin{tabular}{c|c|ccc|c|cc|cc|cc|cc|cc|c}
    \toprule
       Dataset & Distribution     & HeteroFL & DepthFL & FeDepth &  Local  & \multicolumn{2}{c|}{ALexNet} & \multicolumn{2}{c|}{ResNet50} & \multicolumn{2}{c|}{VGG16} & \multicolumn{2}{c|}{DenseNet} & \multicolumn{2}{c|}{MobileNet} & FedStitch \\ 
    \cline{7 - 16}
    & & & & & & FT-Part &FT-Full &FT-Part &FT-Full &FT-Part &FT-Full &FT-Part &FT-Full &FT-Part &FT-Full  \\  
    \midrule
   \multirow{2}{*}{CIFAR10}  &  IID    &  74.15 &	76.23&	79.88&	81.45&	70.10&72.67&	81.32 &  \underline{84.79}  & 78.14 &79.66	 &82.31 &84.52	&73.11 &76.87 &	\textbf{90.49}	      \\
 & Non-IID  & 64.36	&65.68 &	67.24 &	77.55	&65.78 &66.13	&78.89 &78.41 	&74.31 &75.72 &	78.19 & \underline{79.64}  &72.10  &73.00	&\textbf{88.17}	 \\
  \hline
     \multirow{2}{*}{CINIC10}  &  IID    & 59.56	&61.08	&66.67	&64.40	&54.51& 56.77	&65.42 &68.93	&59.38 &62.44 &	65.38  & \underline{69.27}  	&57.26 &60.01	&\textbf{75.66}	      \\
 & Non-IID  & 49.79	&50.39	&53.38	&60.62	&50.94 &52.92	&62.98 &  \underline{63.97} 	&58.21 &60.05	&63.47 &63.82	&55.56 &57.73	&\textbf{71.72} \\
 \hline
     \multirow{2}{*}{CIFAR100}  &  IID    & 32.36	&51.68	&44.24	&45.72 	&43.17 &44.26	&57.56 &57.39	&55.79 &55.54	&56.62  & \underline{57.78} 	&51.13 &53.98 &\textbf{59.89}  \\
     
 & Non-IID  & 25.41	&40.31	&32.72 	&46.19	&39.64 &38.10	&52.02 & \underline{52.87} 	&48.15 &45.04	&47.58 &48.69 &	40.13 &43.73	&\textbf{54.82} \\
  
    \bottomrule     
  \end{tabular} 
  }
  \vspace{-0.5em}
\end{table*}

\section{Evaluation}

% In this section, we first introduce the experimental setup and baselines. Then, we present an empirical evaluation of FedStitch.

\subsection{Experimental Setup}
\subsubsection{Models and Block Pool}
We select five representative neural networks: $alexnet$, $densenet121$, $mobilenet\_v2$, $resnet50$, and $vgg16$. They are pre-trained on the ImageNet-1K \cite{deng2009imagenet} dataset.  
Each pre-trained model is divided into blocks consisting of one or more successive layers, with each convolutional and linear layer having a single input.  Specifically, there are 6 blocks in $alexnet$, 6 blocks in $densenet121$, 20 blocks in $mobilnet\_v2$, 6 blocks in $resnet50$, and 14 blocks in $vgg16$, yielding a block pool of 52 blocks, with 5 starting blocks, 42 intermediate blocks, and 5 terminating blocks. We employ this block pool as the initialization pool in the following experiments.

\subsubsection{Datasets}
We evaluate FedStitch using the following popular datasets including CIFAR10, CIFAR100 \cite{krizhevsky2009learning} and CINIC10 \cite{cinic10}. CIFAR10 has 60k 32x32 color images across 10 classes (6k per class); CIFAR100 includes 60k images in 100 classes (600 per class). CINIC10, blending CIFAR10 and ImageNet, contains 270k images in 10 classes. We aligned ImageNet-1K labels to these datasets and applied the IID/non-IID settings from \textit{Principle 1}.
% For the IID experiment, each dataset is evenly distributed across all clients, with each client having data for all classes. For non-IID, we follow \cite{xu2022fedcorr} to perform data partitioning, for each client, the number of training samples belonging to each class is based on a Dirichlet distribution \cite{frigyik2010introduction} with a concentration parameter set to $\alpha = 1$. 

\subsubsection{Baselines}
 To showcase the effectiveness of FedStitch, we compare it with two types of representative approaches. \textbf{Group 1} includes three methods addressing FL memory constraints through model partitioning: (1)  \textbf{HeteroFL} \cite{diao2020heterofl} prunes the global model via varying model channels; (2) \textbf{DepthFL} \cite{kim2022depthfl}  reducing constraints via mutual self-distillation across classifiers of various depths;
(3) \textbf{FEDEPTH} adaptively decomposes the global model into blocks with different layers and trains blocks sequentially to obtain a full inference model.  
(4) \textbf{Local} uses the same pre-trained models as FedStitch to showcase FL's importance under non-IID conditions. We randomly choose a client to generate the entire stitched network using its local data.
For consistency, all methods use $resnet50$ as the global model, mirroring the structure in their respective studies. For validity, we repeat the same experiment $10$ times with different clients and then average the results. 
\textbf{Group 2} aims to validate our approach against traditional fine-tuning techniques. We fine-tune pre-trained models on local client datasets, considering memory limits and non-IID settings. This includes full model fine-tuning (\textbf{FT-Full}) with limited client involvement, and partial fine-tuning (\textbf{FT-Part}) engaging most clients by adjusting only the model's final layers. Results are then aggregated on the server using FedAvg for comparison.

\begin{figure*}[t]
  \centering
  \captionsetup{font=small}
  \begin{subfigure}[b]{0.16\textwidth}
    \centering
    \includegraphics[width=\textwidth]{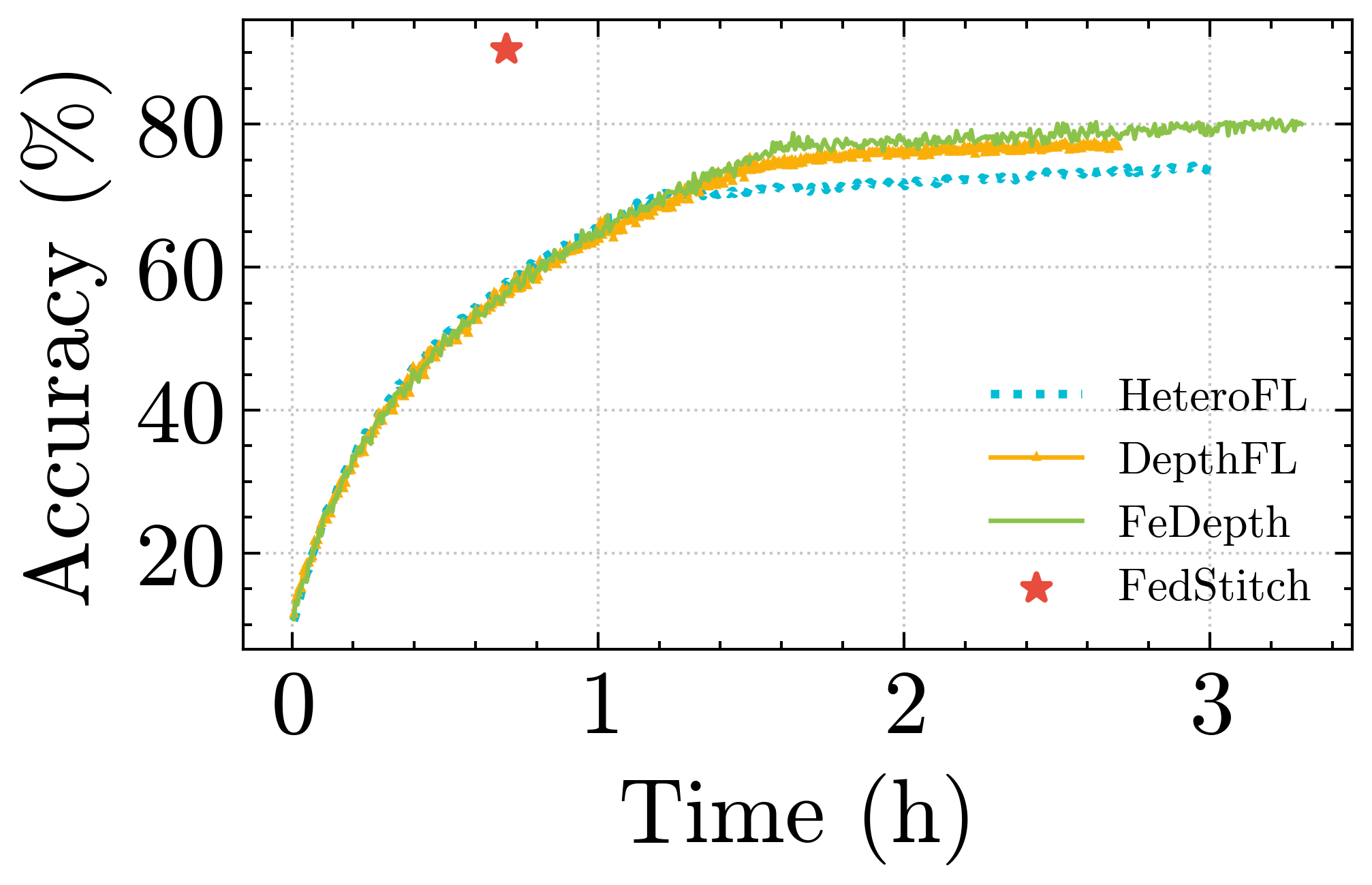}
    \caption{IID (CIFAR10) }
    % \label{cifar10iid}
  \end{subfigure}
  \hfill
  \begin{subfigure}[b]{0.16\textwidth}
    \centering
    \includegraphics[width=\textwidth]{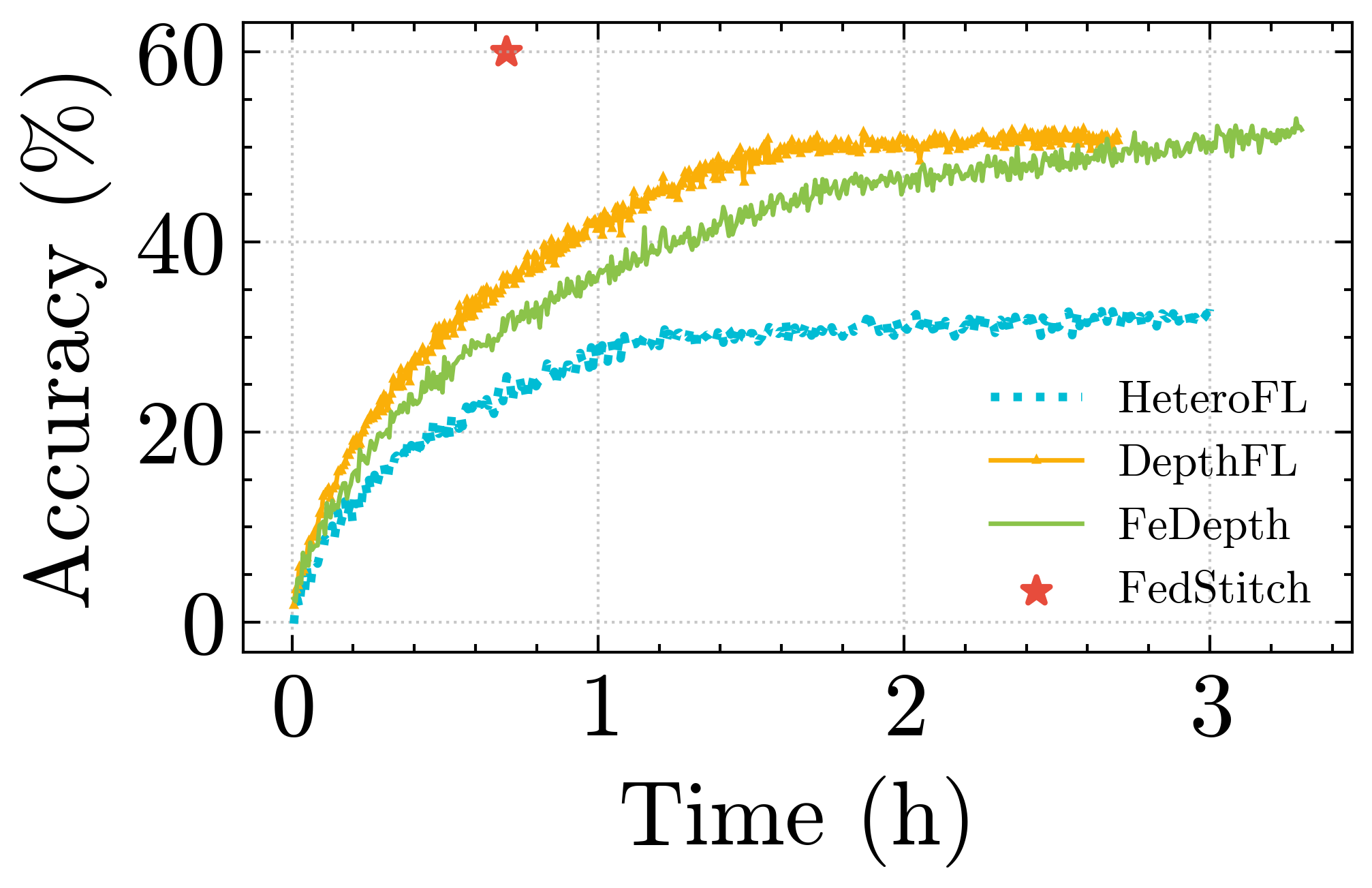}
    \caption{IID (CIFAR100)}
    % \label{cifar10noniid}
  \end{subfigure}
  \hfill
    \begin{subfigure}[b]{0.16\textwidth}
    \centering
    \includegraphics[width=\textwidth]{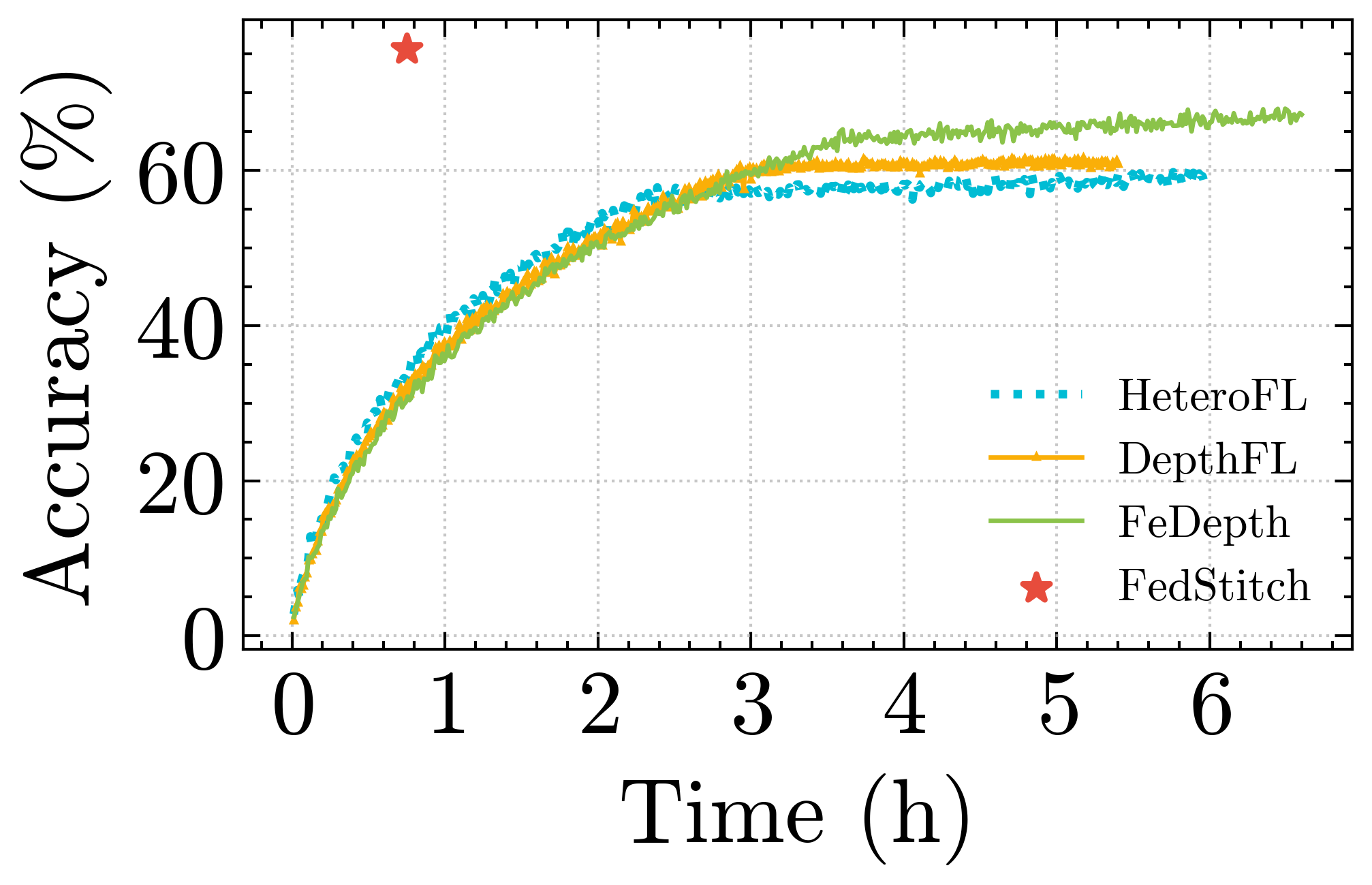}
    \caption{IID (CINIC10)}
    % \label{cifar10noniid}
  \end{subfigure}
 \hfill
      \begin{subfigure}[b]{0.16\textwidth}
    \centering
    \includegraphics[width=\textwidth]{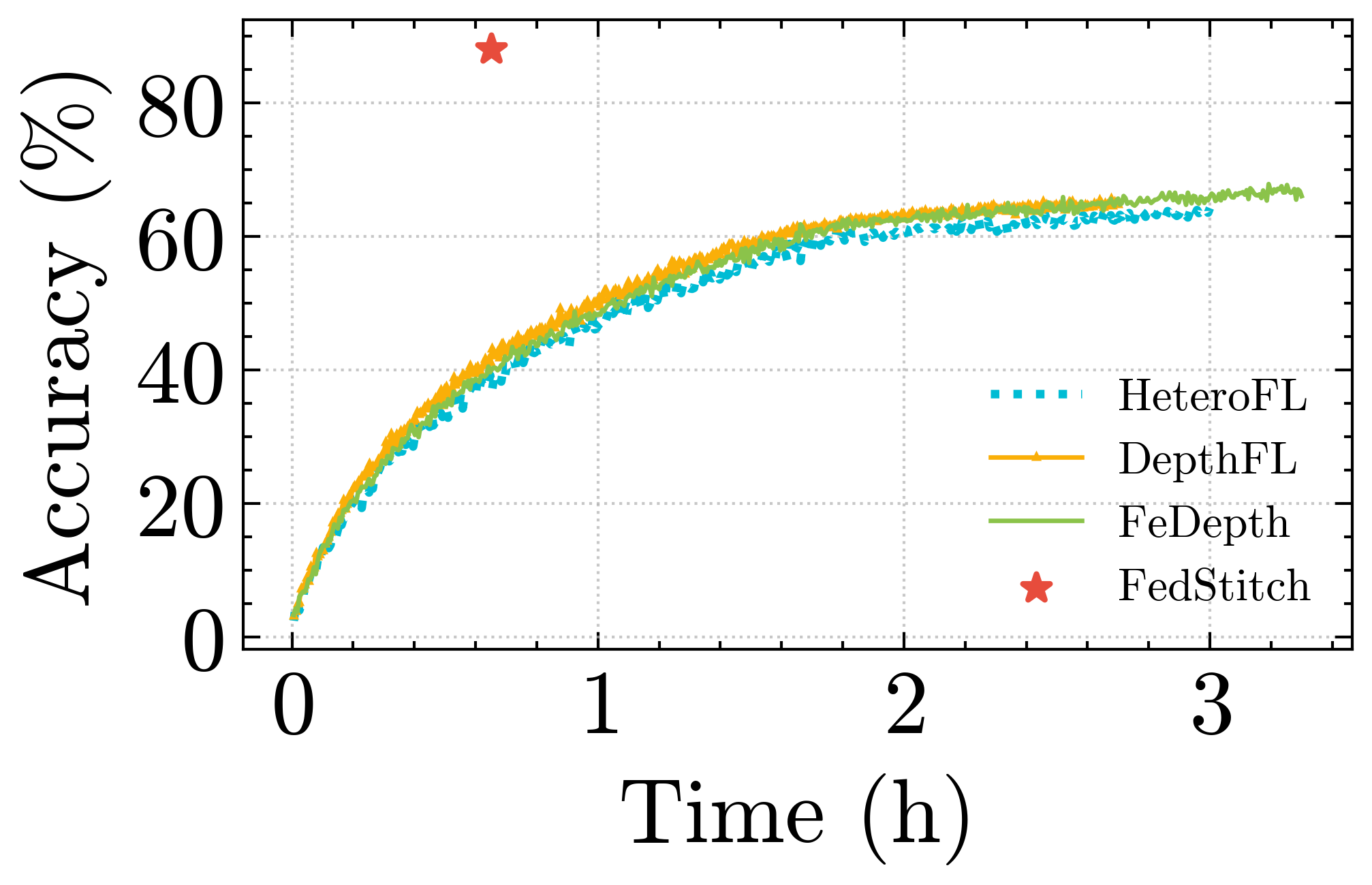}
    \caption{Non-IID (CIFAR10) }
    % \label{cifar10iid}
  \end{subfigure}
  \hfill
  \begin{subfigure}[b]{0.16\textwidth}
    \centering
    \includegraphics[width=\textwidth]{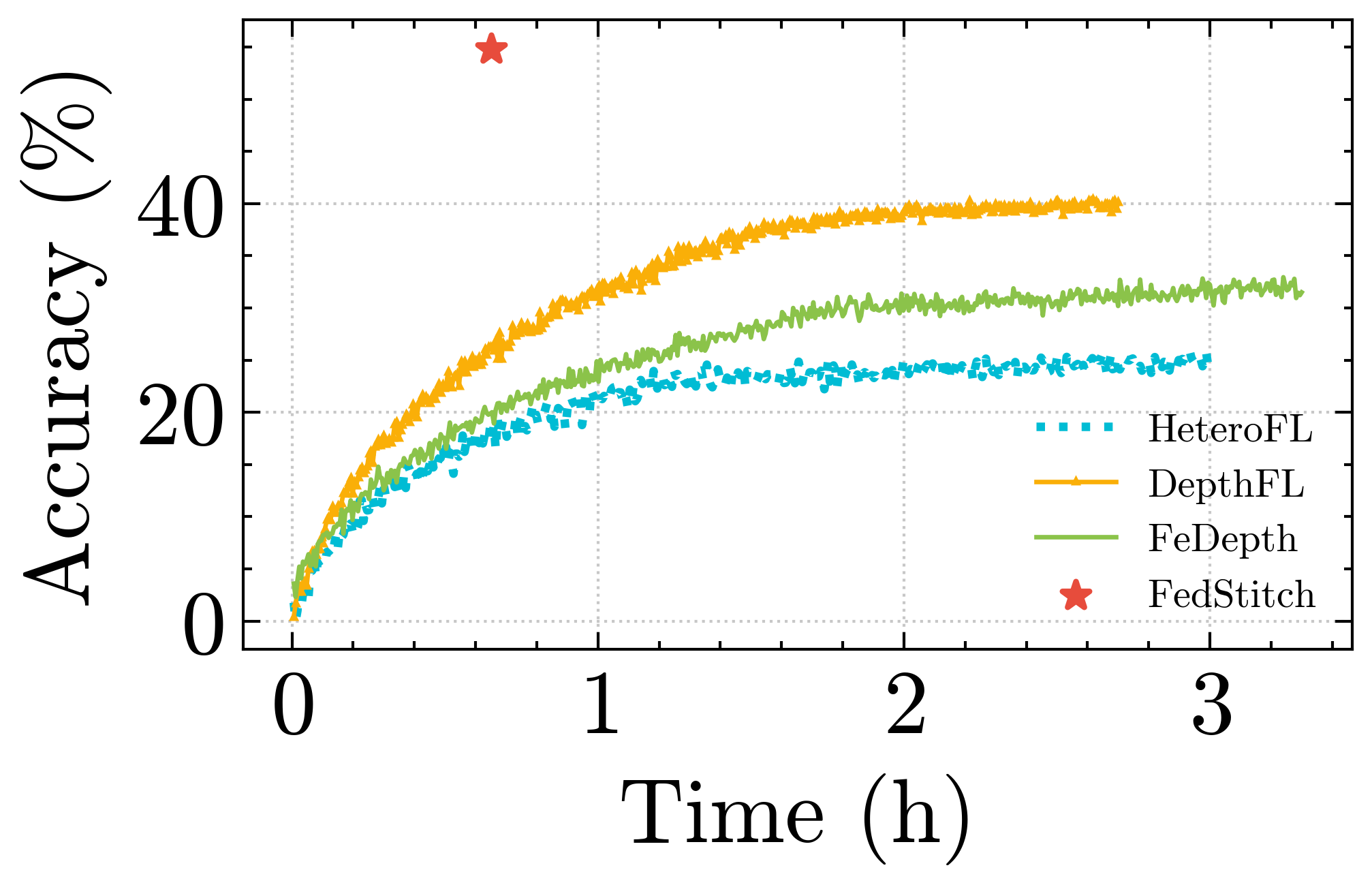}
    \caption{Non-IID (CIFAR100)}
    % \label{cifar10noniid}
  \end{subfigure}
  \hfill
    \begin{subfigure}[b]{0.16\textwidth}
    \centering
    \includegraphics[width=\textwidth]{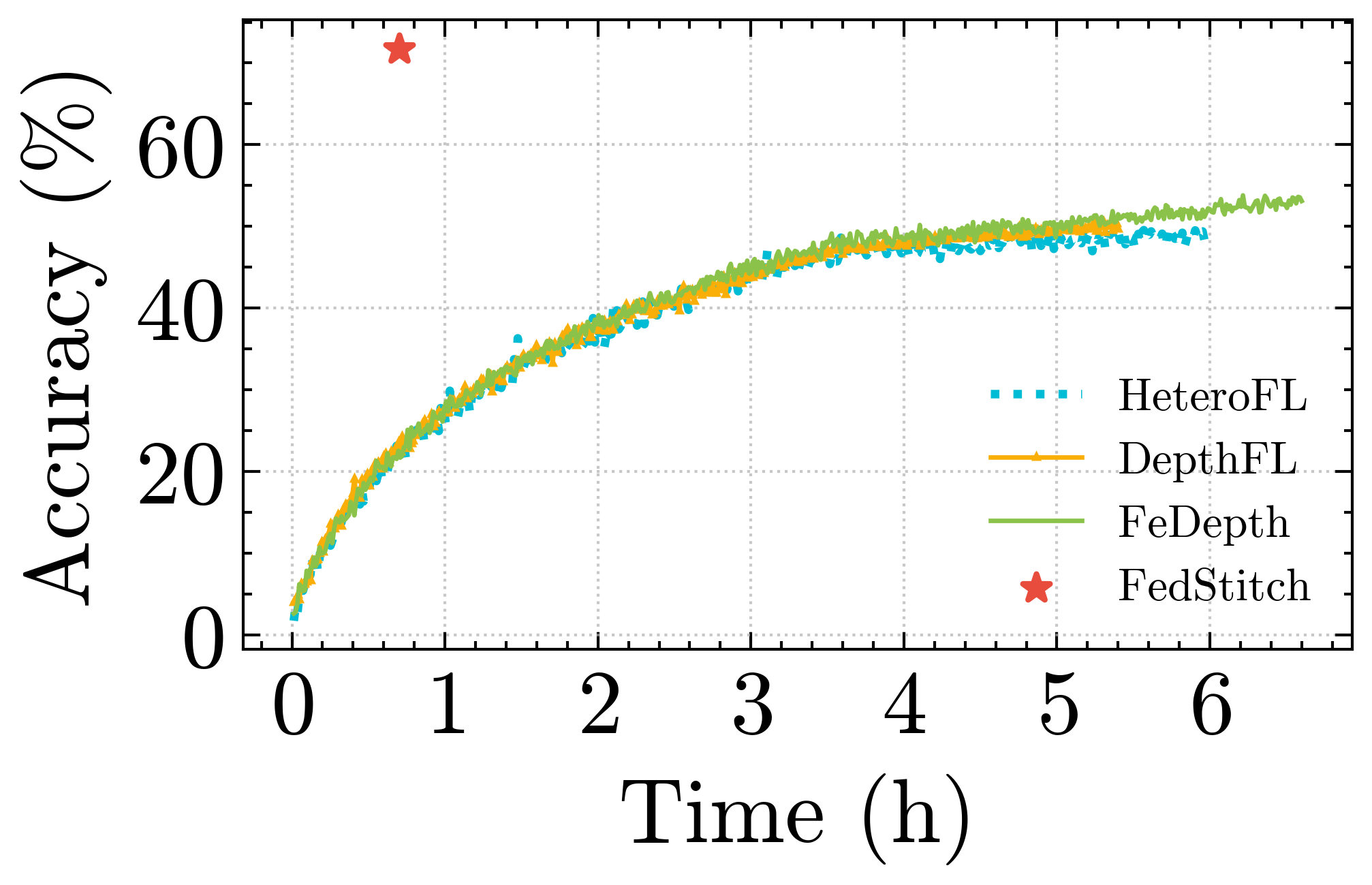}
    \caption{Non-IID(CINIC10)}
    % \label{cifar10noniid}
  \end{subfigure}

    \begin{subfigure}[b]{0.16\textwidth}
    \centering
    \includegraphics[width=\textwidth]{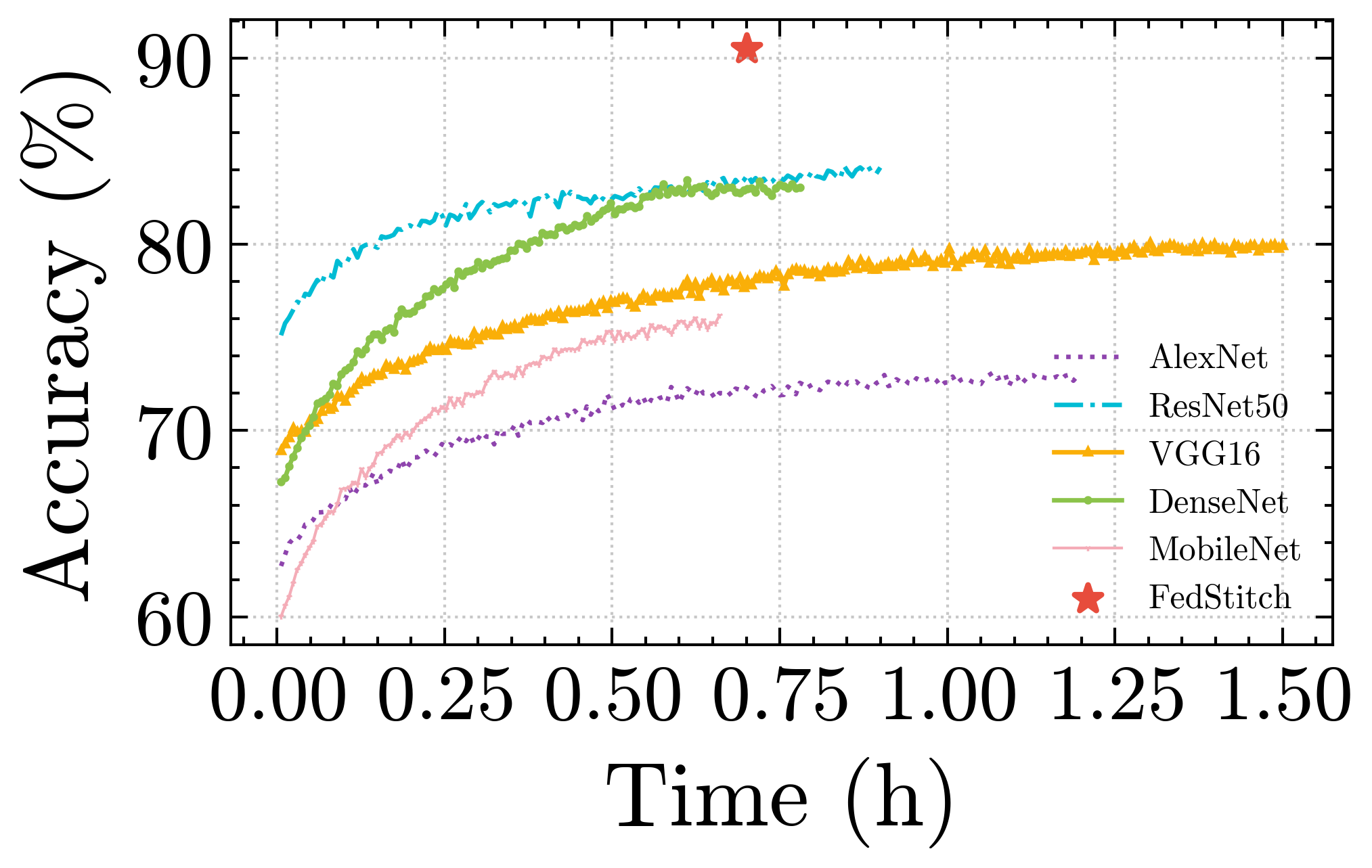}
    \caption{IID (CIFAR10) }
    % \label{cifar10iid}
  \end{subfigure}
  \hfill
  \begin{subfigure}[b]{0.16\textwidth}
    \centering
    \includegraphics[width=\textwidth]{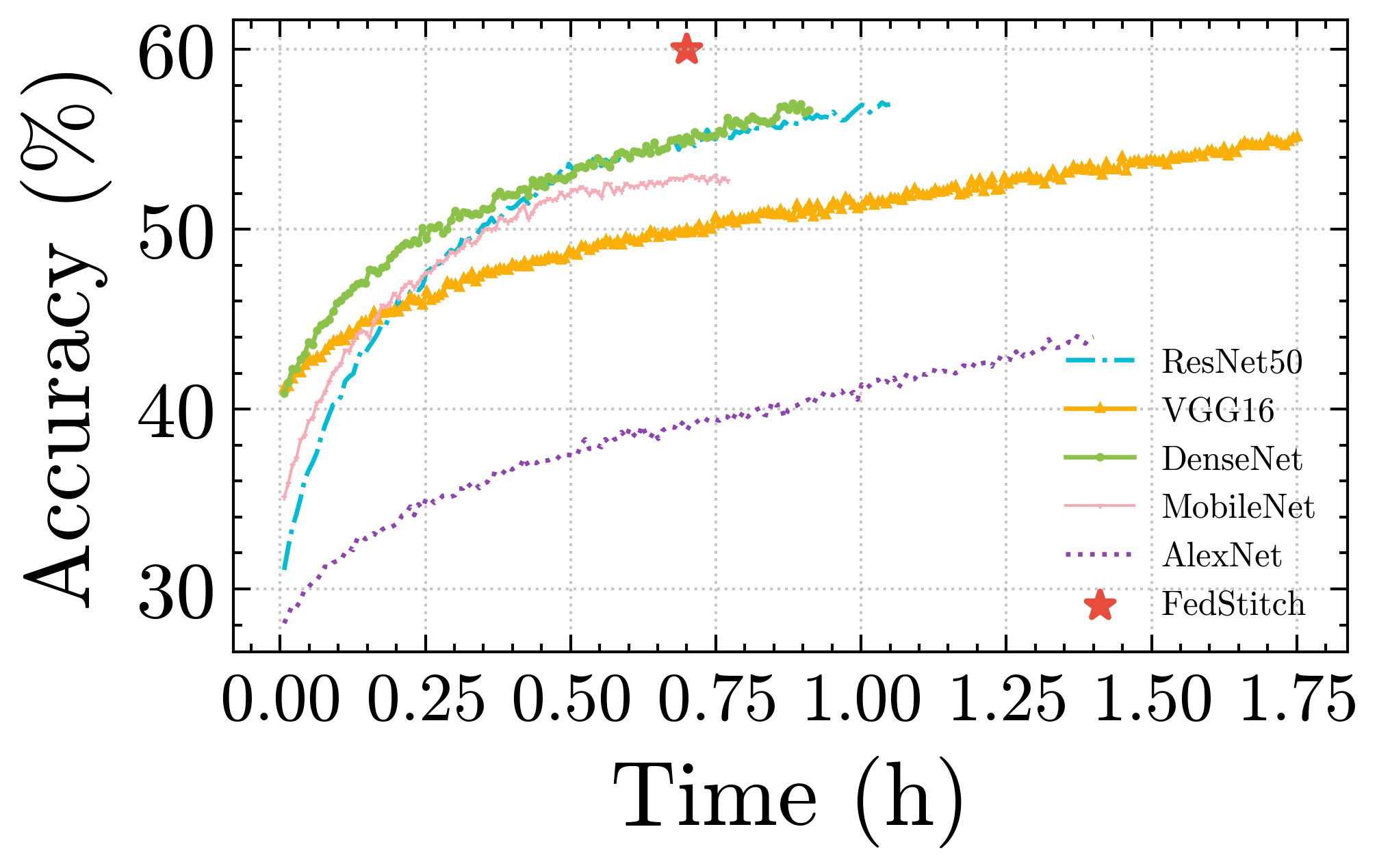}
    \caption{IID (CIFAR100)}
    % \label{cifar10noniid}
  \end{subfigure}
  \hfill
    \begin{subfigure}[b]{0.16\textwidth}
    \centering
    \includegraphics[width=\textwidth]{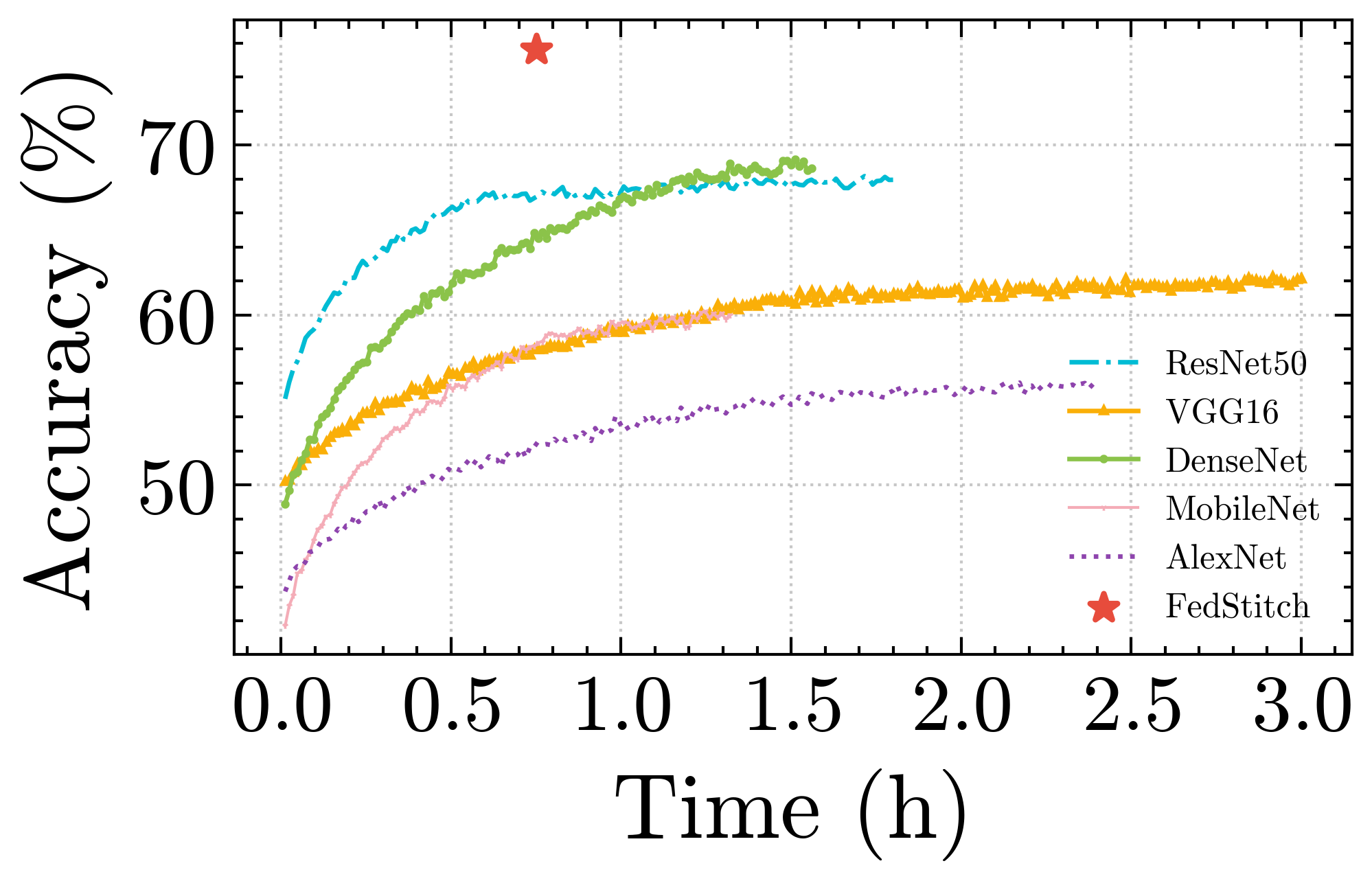}
    \caption{IID (CINIC10)}
    % \label{cifar10noniid}
  \end{subfigure}
 \hfill
      \begin{subfigure}[b]{0.16\textwidth}
    \centering
    \includegraphics[width=\textwidth]{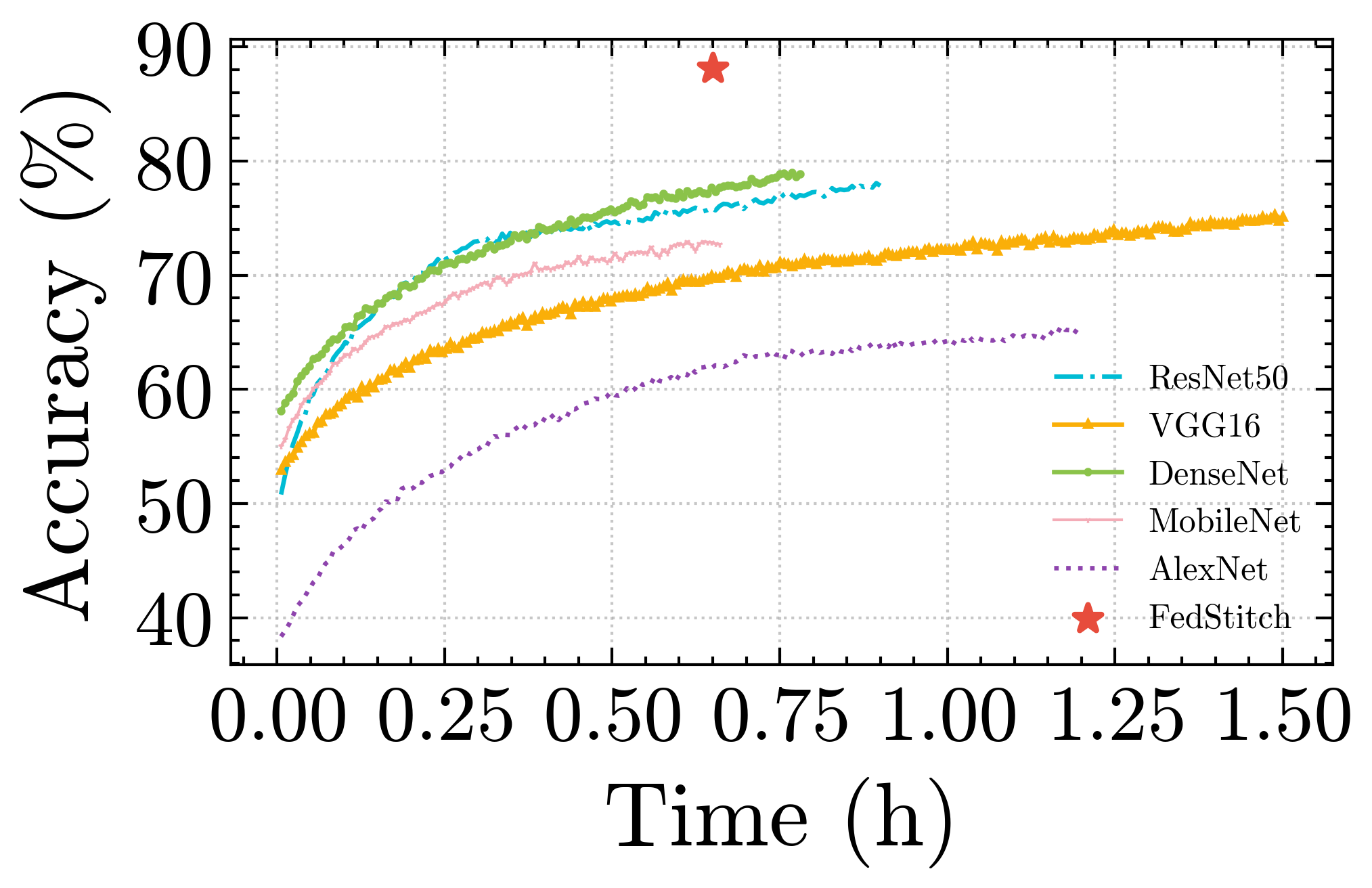}
    \caption{Non-IID (CIFAR10) }
    % \label{cifar10iid}
  \end{subfigure}
  \hfill
  \begin{subfigure}[b]{0.16\textwidth}
    \centering
    \includegraphics[width=\textwidth]{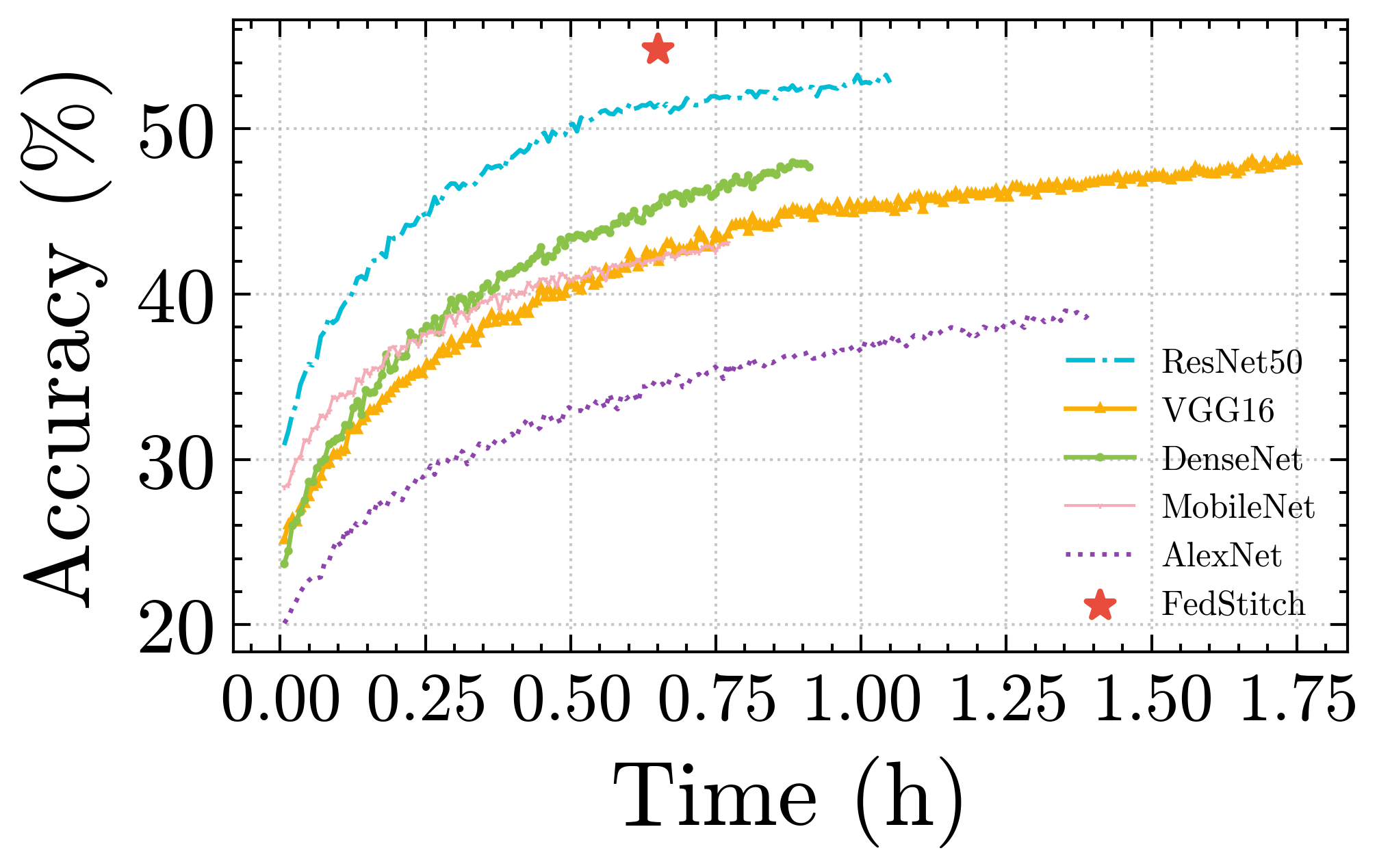}
    \caption{Non-IID (CIFAR100)}
    % \label{cifar10noniid}
  \end{subfigure}
  \hfill
    \begin{subfigure}[b]{0.16\textwidth}
    \centering
    \includegraphics[width=\textwidth]{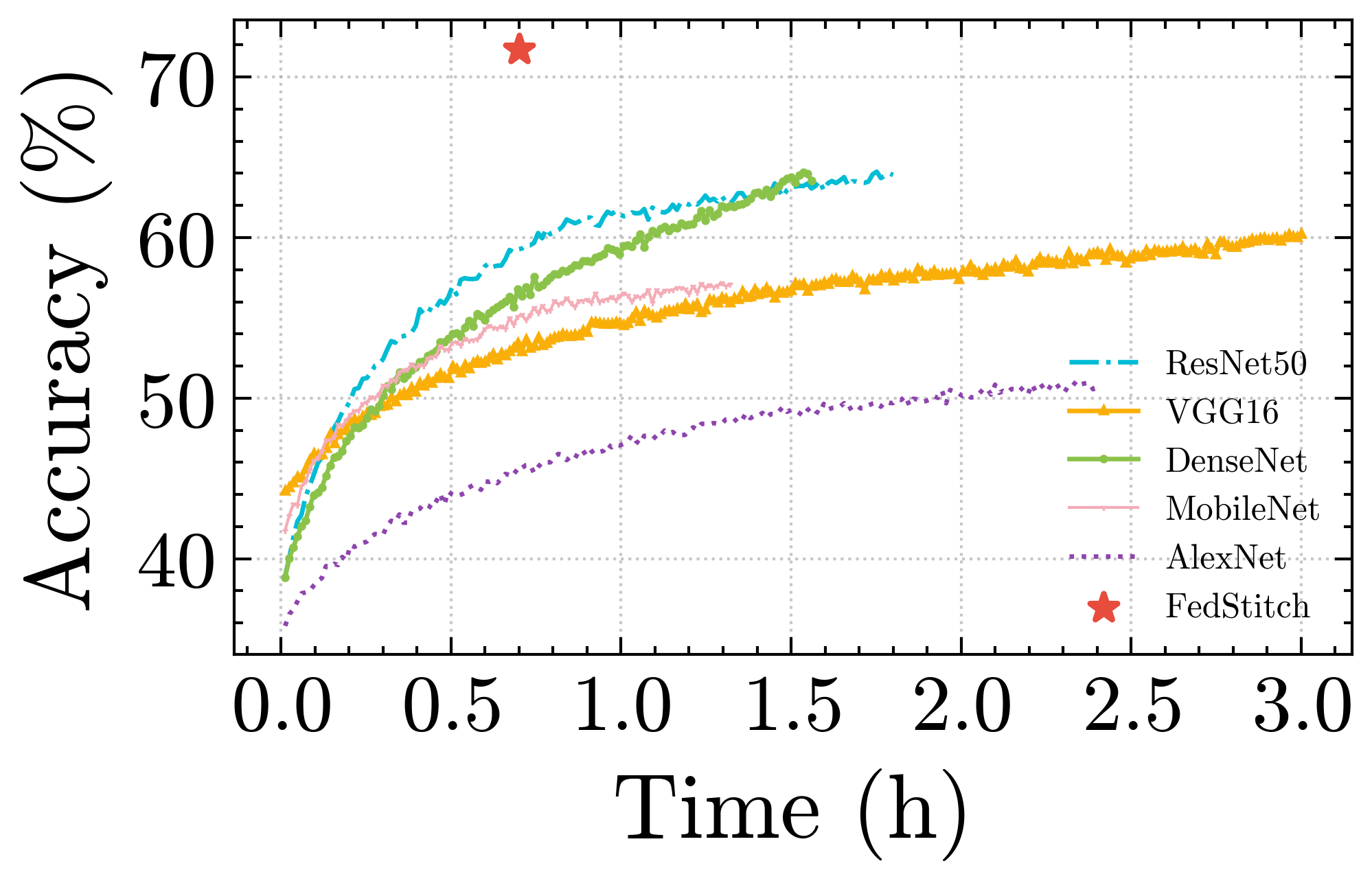}
    \caption{Non-IID (CINIC10)}
    % \label{cifar10noniid}
  \end{subfigure}

  \caption{ Efficiency comparison of various schemes with \textbf{baseline group 1} (a-f) and \textbf{baseline group 2} (g-l) on CIFAR10, CINIC10, and CIFAR100 datasets in IID/Non-IID scenarios on Jetson TX2. The performance of FedStitch is denoted as \textcolor{red}{$\bigstar$}. $alexnet$, $resnet50$, $vgg16$, $densenet$, and $mobilenet$ refer to one of the two fine-tuning methods (\textbf{FT-Full} and \textbf{FT-Part}) that yielded higher accuracy. }
   \label{speed up1}
   \vspace{-1.5em}
\end{figure*}

\subsubsection{Implementation Details}

We choose Nvidia Jetson TX2 as the embedded device for deploying the on-device FL system. We use the Monsoon Power Monitor to record energy consumption and htop to monitor memory usage. 
We configured 100 users, with 10 users participating in each round. 
The model sizes of 5 pre-trained models are 94MB ($resnet50$), 33MB ($densenet121$), 14MB ($mobilenet\_v2$),  228MB ($alexnet$), and 537MB ($vgg16$). Thus total memory of 906MB is needed to store these pre-trained models.  We split the users into 4 groups, with each group representing 30\%, 30\%, 30\%, and 10\% of the total. The memory constraints for each group are 1GB, 2GB, 4GB, and 8GB, respectively. Under this setting, 
 three baselines in group 1 train a global model individually using their respective model partition and training methods based on users' memory budgets.
Compared to the model size, the memory overhead of inference can be negligible. Therefore, in this memory setting, even users with the smallest memory allocation can participate in the entire process of FedStitch.  In real-world scenarios, even if there are devices with highly limited resources that cannot participate in local block selection,  we can still ensure that the majority of devices contribute to the network generation process.

For the baselines in group 1, we adopt SGD as the optimizer with a momentum of 0.9, weight decay of 0.0005, and a learning rate of 0.01. We set the number of local epochs as 10, the local batch size as 128, and the number of global epochs as 500.
For the baselines in group 2, we adjust the learning rate to 0.001 and global epochs to 100. 
For FedStitch, the process of choosing the next block for the current candidate stitched networks is repeated for 5 rounds, and the overall results are aggregated to determine the optimal block. The number of block selections $K$ is 3.
The size of batch samples used for CKA computation is 64. The total epoch of stitching is usually less than 100 depending on the depths of generated stitched networks, and each epoch consumes much less time than baselines due to avoided training.

\subsection{Performance Effectiveness}
Through the method in Section \ref{rl}, a total number of 22 stitched neural networks are generated.
% Table \ref{tab:statistics} summarizes the statistics of the top-5 generated networks in non-IID on CIFAR10, including the test accuracy, CKA score, and model size. We find that although a higher CKA score generally corresponds to better network performance, the model with the best performance may not necessarily have the highest CKA score. This observation inspires us to select multiple blocks for stitching in each round of the generation process. 
% Most well-performance stitched networks have smaller sizes compared to pre-trained models.
We choose the generated networks with the best performance for comparison. Table \ref{tab:result} demonstrates the comparison results with baselines in all datasets.  We find that (1) FedStitch outperforms the best baseline in group 1 that addresses memory constraints in both IID and non-IID in all datasets achieving \textbf{10.61\%} and \textbf{20.93\%} absolute improvements respectively in CIFAR10. There are two primary reasons for this improvement. Firstly, it eliminates the need for training entirely, significantly reducing memory requirements. This allows users previously excluded due to memory constraints to contribute to the global model. Secondly, compared to training from scratch, FedStitch leverages pre-trained knowledge from public datasets. Consequently, the generated model is more robust, and can effectively avoid overfitting, exhibiting enhanced generalization and adaptability to new tasks. 
(2) Compared to the traditional fine-tuning on pre-trained models in group 2, our approach achieves superior performance on new tasks in three datasets achieving \textbf{5.7\%} and \textbf{8.53\%} absolute improvements for IID and non-IID in CIFAR10. Due to memory constraints, only a few users can fine-tune the entire model, resulting in a sub-optimal global model. Fine-tuning only the last few layers of the network, on the other hand, is influenced significantly by the data heterogeneity among users, affecting the performance of the aggregated global model. Our method, through block selecting and stitching, enables the participation of most users and makes the generated model enhanced adaptability to new tasks. Through the design in Section \ref{rl}, we significantly mitigate the impact of data heterogeneity. Consequently, utilizing the same pre-trained models, our approach outperforms traditional fine-tuning methods on new tasks. 
(3) Compared to stitched network on a single device, FedStitch has better performance achieving \textbf{9.04\%} and \textbf{10.62\%} absolute improvements in CIFAR10, because the data distribution in FL is highly skewed, the models generated by users lacking some classes have inferior expressive capability on new tasks.

\subsection{System Efficiency}  

By stitching the neural network instead of training from scratch, FedStitch improves the model effectiveness and system efficiency at the same time. In this section, we evaluate our system performance from three perspectives: memory overhead reduction, computation speed-up, and energy consumption optimization.

\begin{figure}[t]
  \centering
    \includegraphics[width=0.24\textwidth]{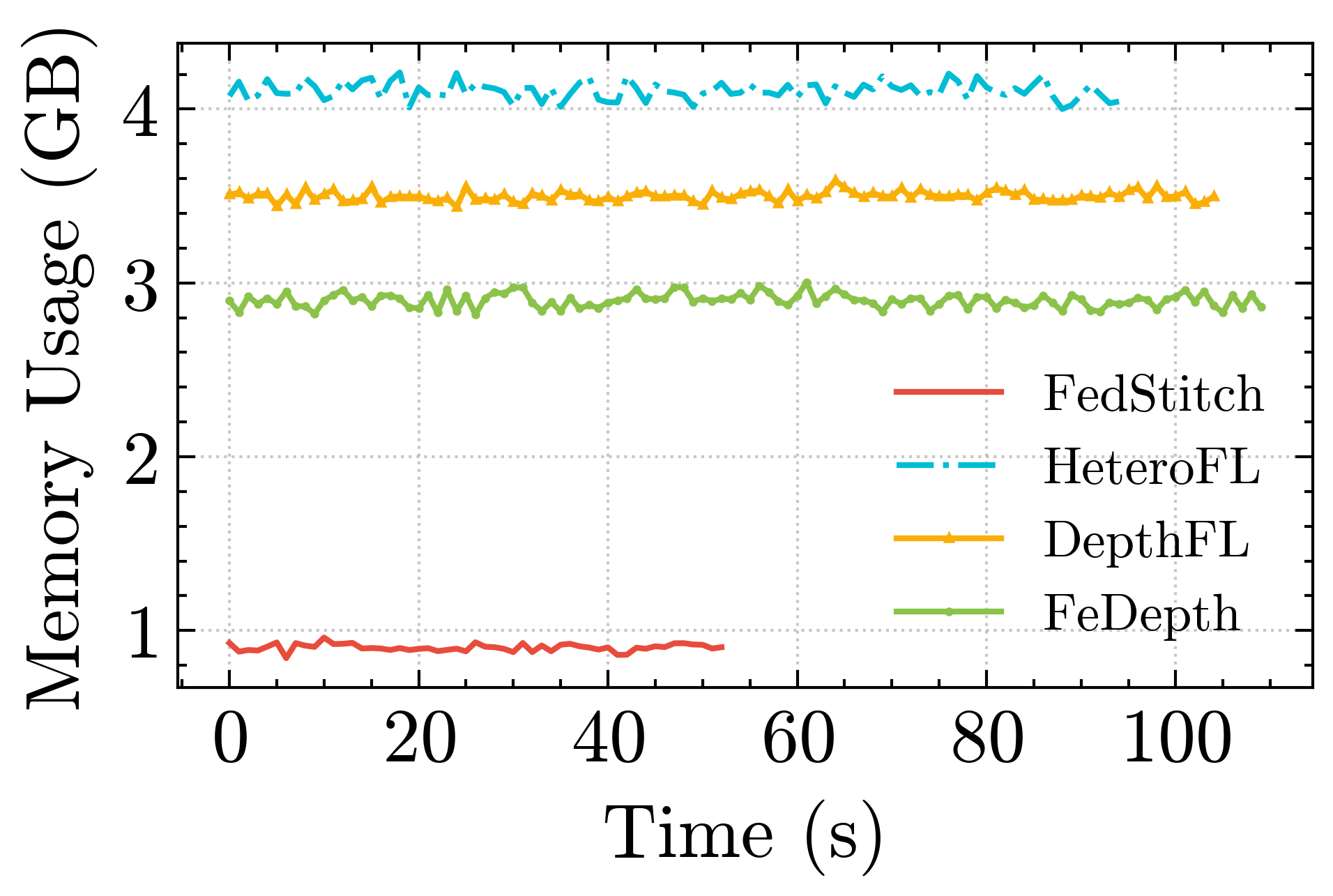}
    \includegraphics[width=0.24\textwidth]{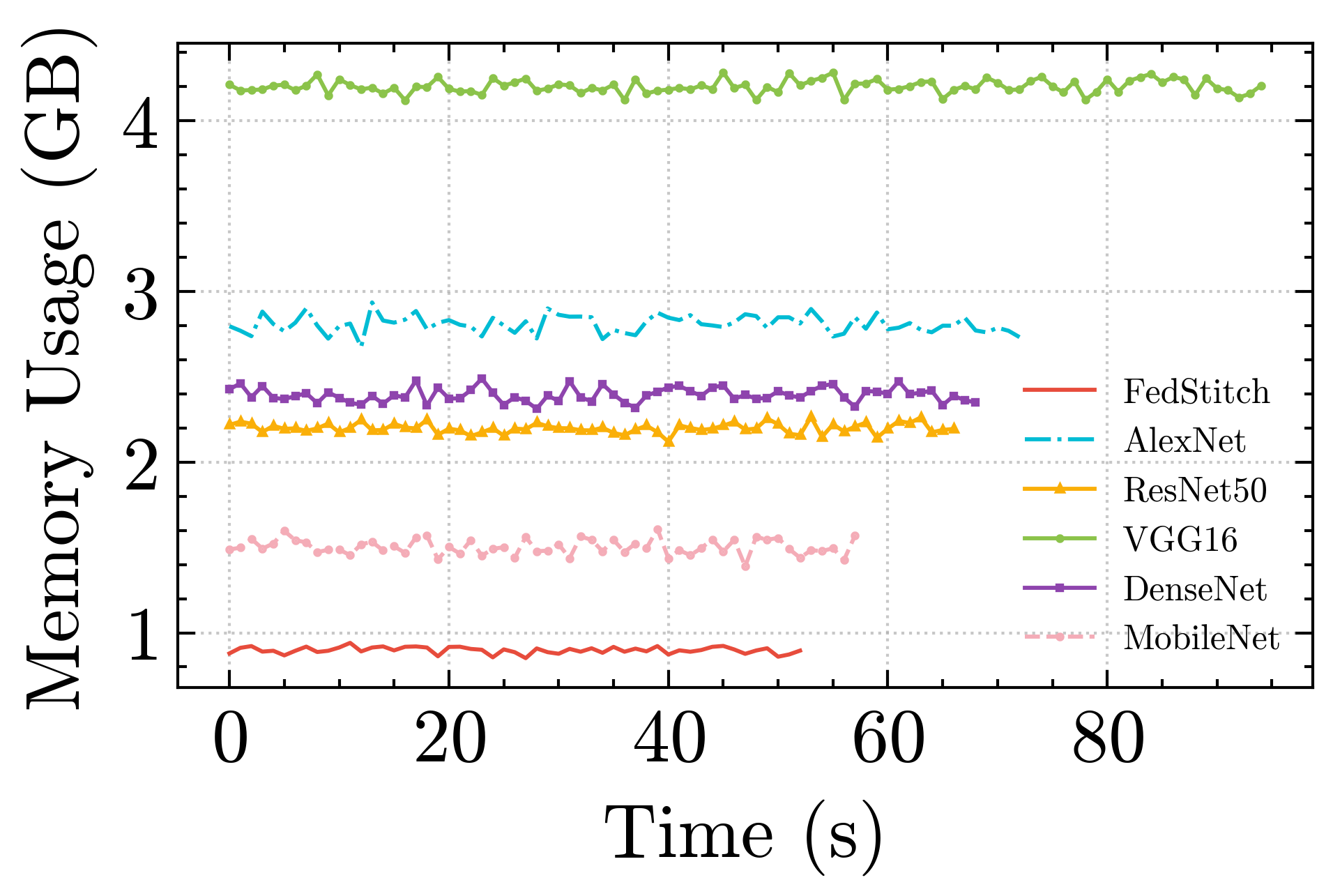}
 \caption{Memory consumption per round. Left: group 1; Right: group 2.}
 \label{memory-overall}
 \vspace{-1.5em}
\end{figure}

\subsubsection{Training Speed-up}  To validate the efficiency of FedStitch, we evaluate its acceleration effect for computation on all datasets in both IID and non-IID scenarios. 
Fig. \ref{speed up1} shows the time-accuracy results with different baseline groups, where the X-axis represents time (hours) and the Y-axis represents the related testing accuracy.  In the IID scenario,  compared to both groups of baselines, whether training from scratch or fine-tuning the pre-trained model, FedStitch significantly reduces the overall time by continuously reducing the size of the candidate block pool and avoiding the training process, which eliminates the computation of backpropagation, and maintains a comparable accuracy. For the baseline group training from scratch, FedStitch yields a speedup of up to 5.02$\times$, 7.34$\times$, and 6.35$\times$ on CIFAR10, CINIC10, and CIFAR100 datasets. In the non-IID scenario, thanks to the RL-based weighted aggregation design, FedStitch does not suffer a significant accuracy drop compared to IID and achieves up to 6.01$\times$, 8.12$\times$, and 5.42$\times$ speedup on three datasets.

\subsubsection{Memory Overhead Reduction}
We evaluate the memory overhead of FedStitch on Jetson TX2, using htop to monitor the memory usage during the generation process. Fig. \ref{memory-overall} compares the memory usage in one round with two baseline groups. 
From Fig. \ref{memory-overall}, it can be observed that, due to the elimination of the need for training, a significant amount of memory overhead, such as activation in backpropagation,  can be saved. As a result, compared to the baseline, the memory requirements in FedStitch can be reduced by 41.2-79.5\%. During the process of generating the stitched network using 5 pre-trained networks, the maximum memory consumption typically does not exceed 1GB. This allows FedStitch to be deployed on almost all edge devices, thereby addressing the challenges of memory constraints in FL.
% 要强调 因为我们需要的sample始终不变 所以需要的时间不受数据集大小的影响

\subsubsection{Energy Consumption Optimization}
\begin{figure}[t]
  \centering
    \includegraphics[width=0.24\textwidth]{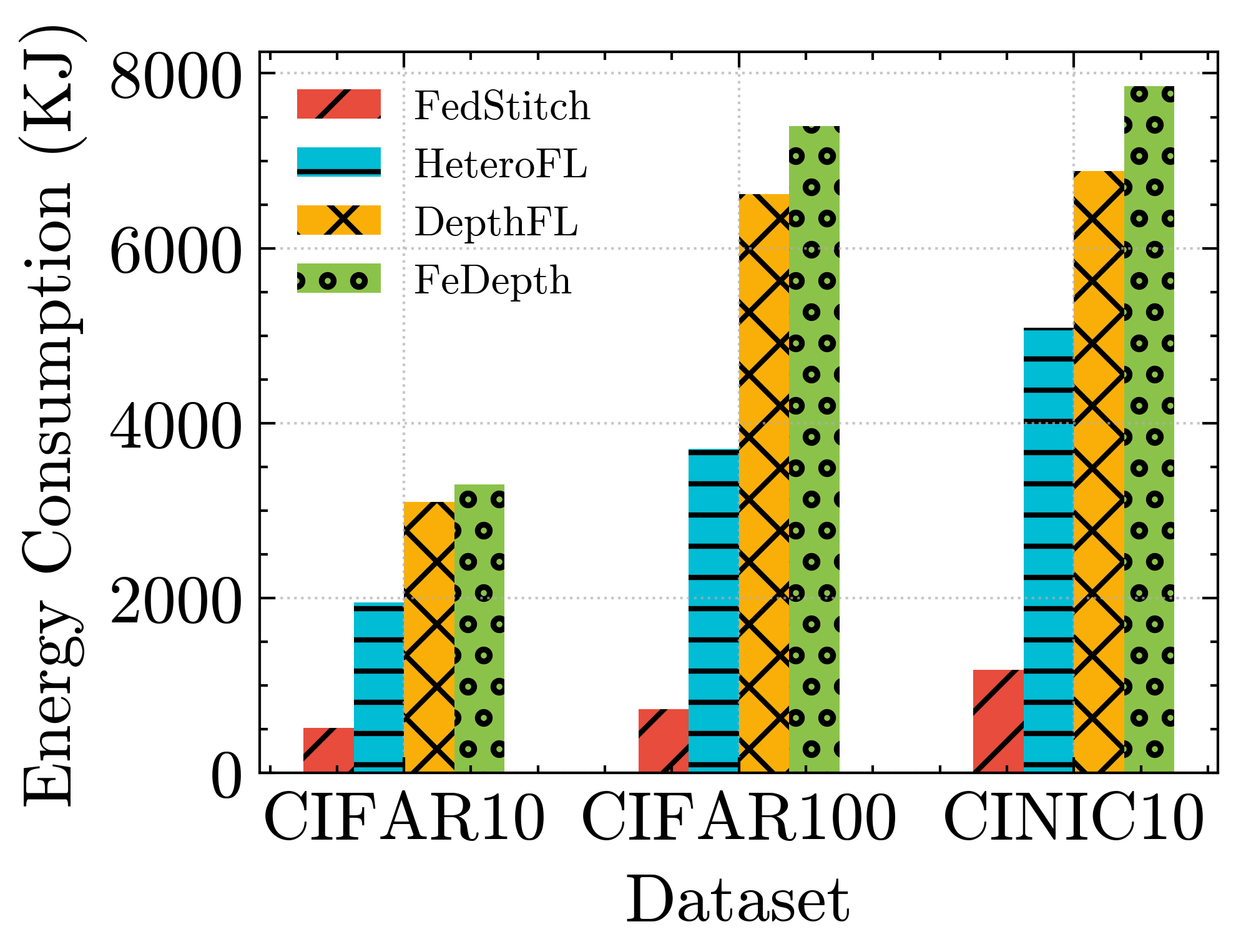}
    \includegraphics[width=0.24\textwidth]{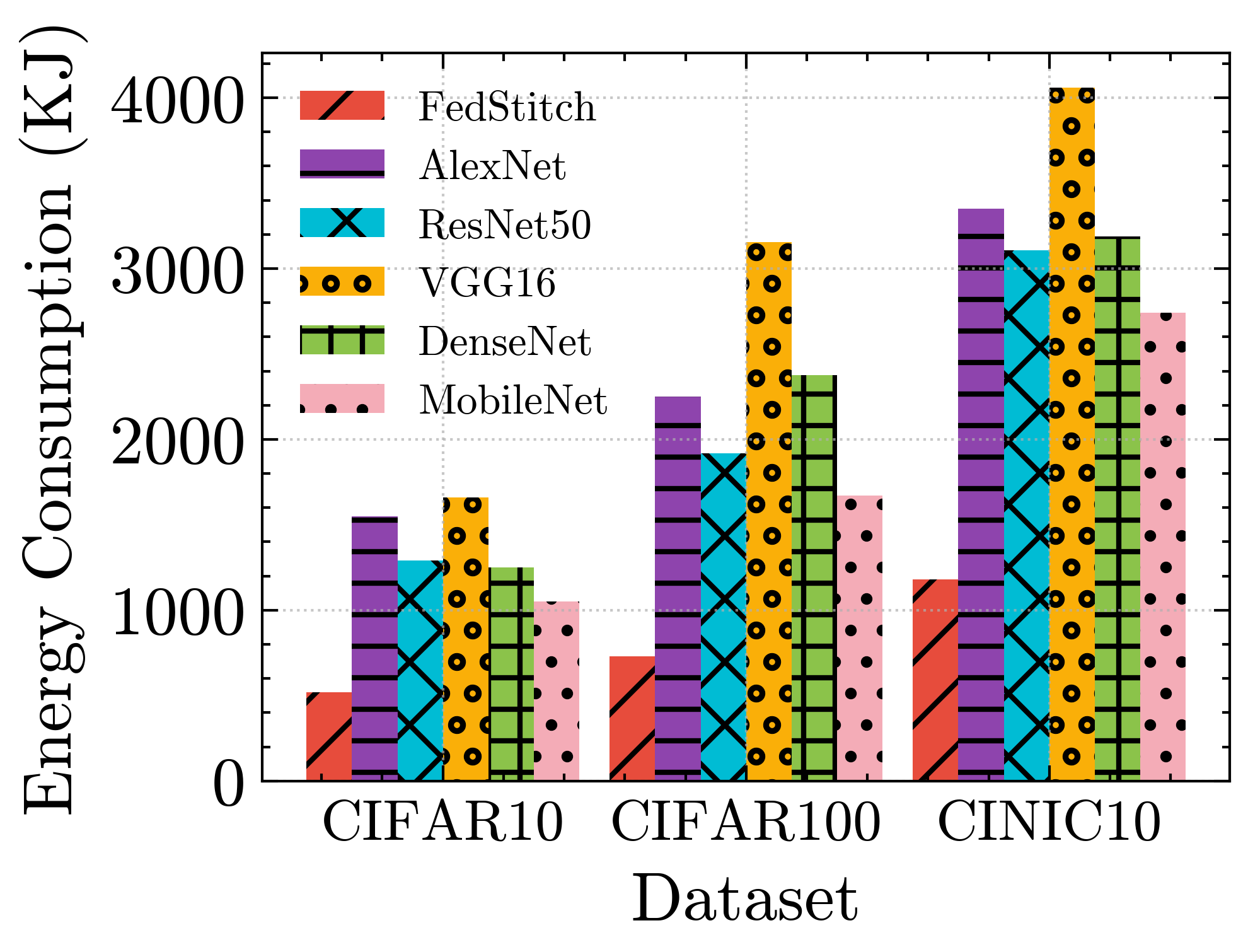}
  \caption{ Eenergy consumption on CIFAR10, CIFAR100, and CINIC10 datasets (IID) on Jetson TX2. Left: group 1; Right: group 2.}
  \label{energy}
  \vspace{-1.1em}
\end{figure}

To assess the efficiency of power consumption, we evaluated the performance of FedStitch on CIFAR10, CIFAR100, and CINIC10 datasets, using Monsoon Power Monitor on Jetson TX2. Fig. \ref{energy} demonstrates the comparison results with two groups of baseline, where the X-axis represents the various schemes of different datasets and the Y-axis represents the related energy consumption (kilojoule, KJ).  
Compared to the baselines, FedStitch exhibits significant reductions in both computational overhead and aggregation time. This is attributed to the fact that FedStitch eliminates the need for training-related computations and also reduces a considerable amount of inference-related computations due to the search space optimizer. Furthermore, our design of the local energy coordinator allows each client to run the block selection process with minimal energy consumption configuration while being able to catch the deadline, instead of the highest frequency execution as with default DVFS. This leads to further reductions in overall energy consumption. There is a similar trend in the non-IID scenario. In summary, FedStitch achieves energy-saving up to 89.41\%. 

\subsection{Ablation Study}

To explore the impact of the different modules of FedStitch, we conduct the following ablation experiments.

\subsubsection{ Effectiveness of On-the-fly Search Space Reduction}

\begin{figure}[t]
  \centering 
    \includegraphics[width=0.22\textwidth]{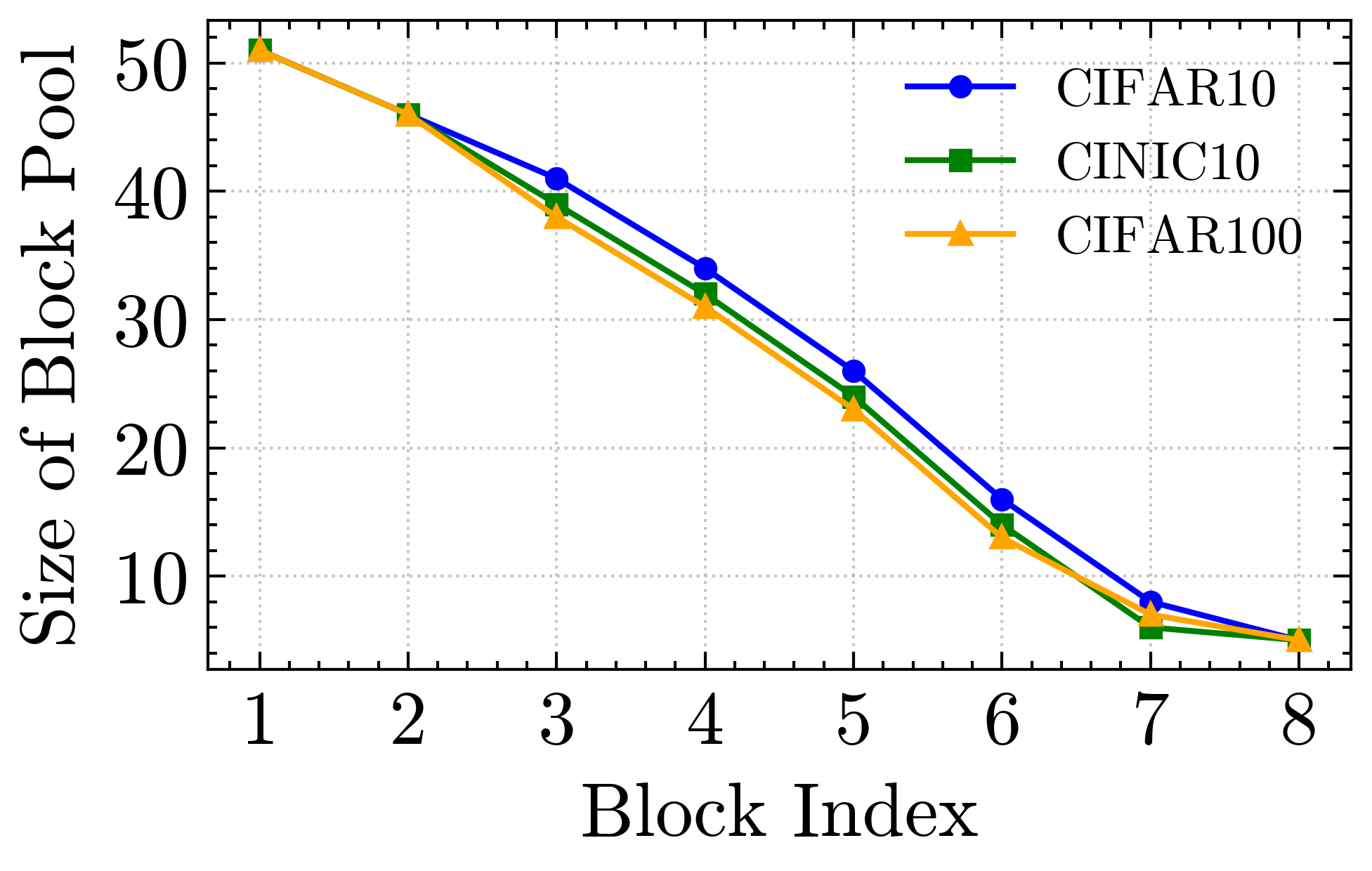}
    \includegraphics[width=0.22\textwidth]{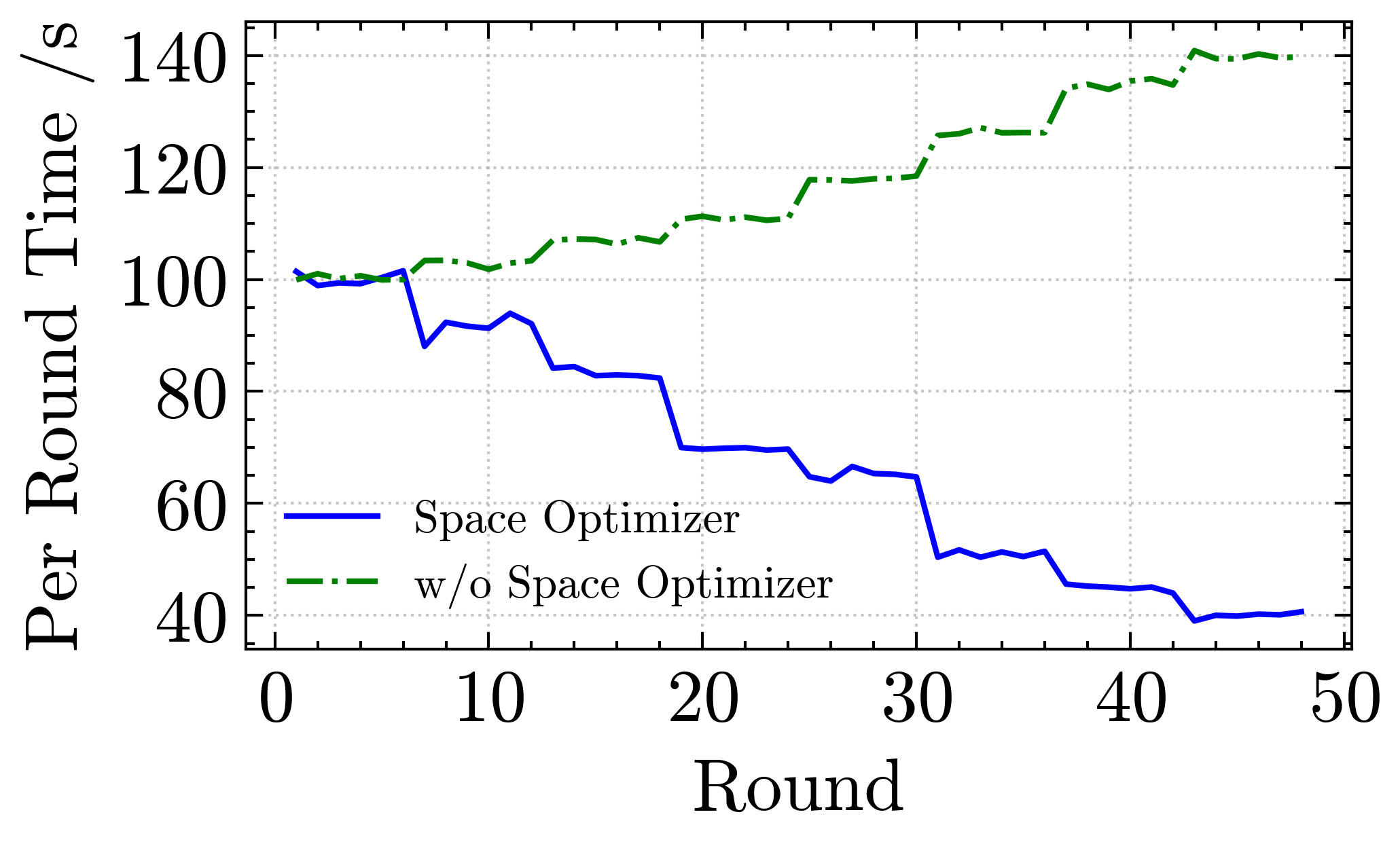}
  \caption{Performance breakdown of search space optimizer. Left: Search space decreases during the stitched network generation. Right: Per round time.}
   \label{searchspace}
   \vspace{-1.5em}
\end{figure}

To illustrate the effectiveness of our design, we take an 8-block generated stitched network as an example. 
As shown in Fig. \ref{searchspace} (left), given 5 pre-trained models,  there are more than 50 blocks in the 
initial block pool. During the network generation process, the selection of the next block is accompanied by a continuous decrease in the size of the candidate block pool. 
The pool size is continuously reduced to less than 10 for the last block selection.   In contrast, without the space optimizer, the block pool would remain at its initial size ($>50$) throughout the entire process. 
From Fig. \ref{searchspace}, 
by continuously reducing the size of the candidate block pool during the stitching process, the time of block selection in each round decreases. Without employing the space optimizer, as the candidate stitched network goes deeper, the related inference time increases. Therefore, in the scenario where the size of the block pool remains unchanged, the block selection in each round will take more time. Overall,  FedStitch achieved a speedup of approximately $42.1\%$ compared to the one without the space optimizer.
Therefore, our space optimizer effectively reduces the computational complexity during network generation, thereby accelerating the entire process.

\subsubsection{ Effectiveness of RL-based Weighted Aggregation}
 We conducted a noise experiment using the traditional FedAvg method for block aggregation in server. From Table \ref{ab: RL},  we observed accuracy drops by 3.52\%, 7.7\%, and 5.49\% over CIFAR10, CIFAR100, and CINIC10 datasets. This indicates that only relying on the averaging aggregation to filter blocks from different users with various non-IID level data is not robust in the heterogeneous data distribution. Our method effectively identifies the user with lower-level non-IID data and assigns them with higher weight, helping to improve the performance of stitched networks. 

\begin{table}
\centering
    \caption{Performance comparison (top-1 accuracy) with and without RL weighted aggregator in non-IID.}

        \begin{tabular}{c|ccc}

    \toprule
 Experient &  CIFAR10 & CIFAR100 & CINIC10  \\      
    \midrule
 FedAvg &   84.65 &   47.12 & 66.23 \\
 
 FedStitch  & \textbf{88.17} & \textbf{54.82} & \textbf{71.72} \\
    \bottomrule     
  \end{tabular} 
      \label{ab: RL}
      \vspace{-2em}
\end{table}

\begin{figure}[!ht]
\centerline{\includegraphics[width=0.6\linewidth]{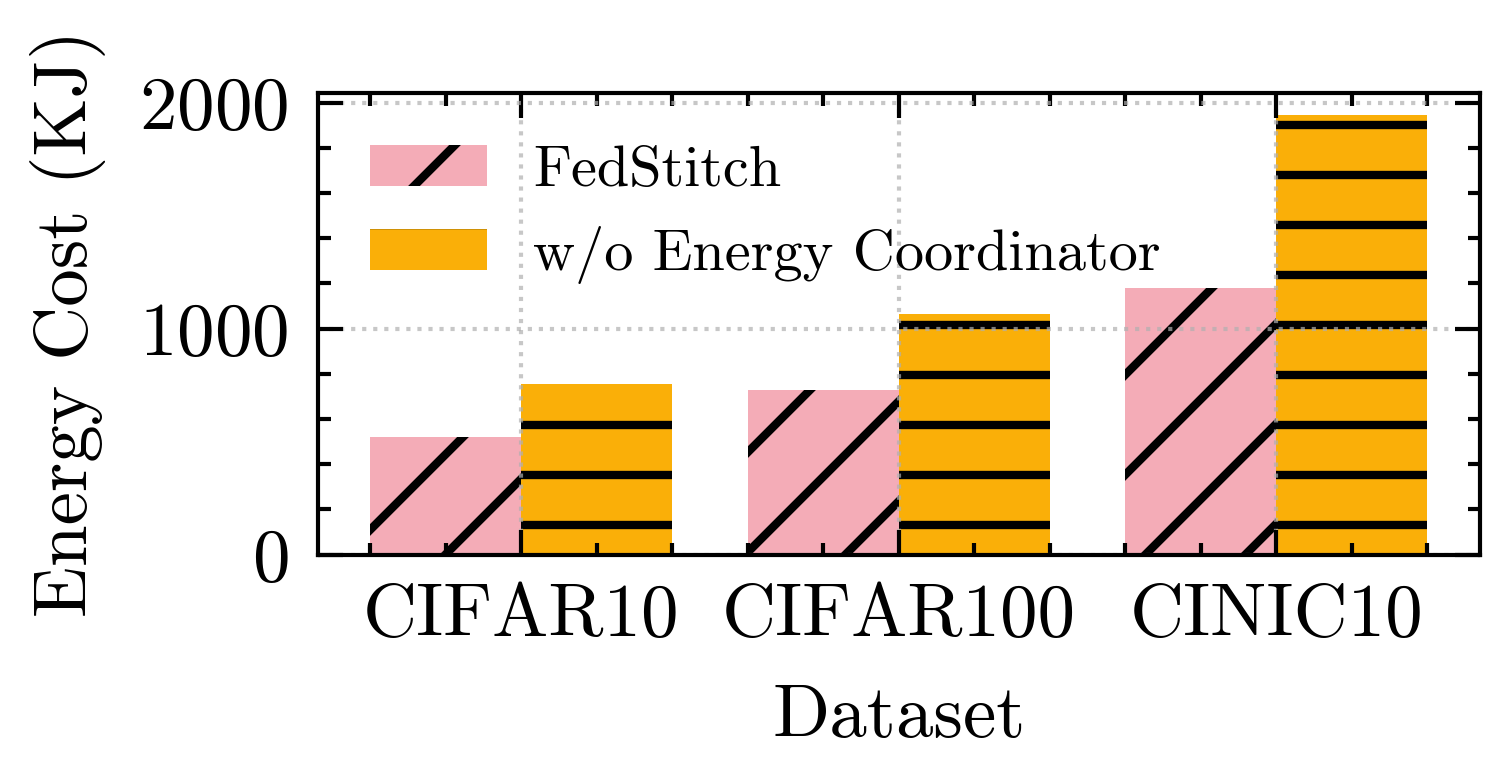}}
\caption{Effectiveness of local energy coordinator.}
\label{energy_ablation}
\vspace{-1em}
\end{figure}

\subsubsection{ Effectiveness of Local Energy Coordinator}
We design a noise experiment, which constantly selects the configuration with the highest frequency and voltage for each client.  
From Fig. \ref{energy_ablation}, FedStitch acheives energy reductions by 32.3\%, 30.4\%, 40.1\% on CIFAR10, CIFAR100, and CINIC10 datasets. It indicates 
FedStitch effectively lowers energy consumption on edge devices while maintaining aggregation time per round, thus preserving the performance of stitched networks.

\section{Conclusion}

% In this paper,  we propose FedStitch, a novel FL paradigm that addresses the challenge of memory constraints and statistical heterogeneity.  FedStitch utilizes the pre-trained networks as auxiliaries, splits them into blocks, and these blocks are stitched into new stitched networks on downstream task. Compared to training a network from scratch or fine-tuning the pre-trained networks, our method requires no training and only minimal data. This significantly saves computational resources and the memory space used for
% storing intermediate outputs and gradients during the backpropagation process.  For the performance drop due to the statistical heterogeneity in FL, we propose a reinforcement learning-based weighted aggregation algorithm with cross-validation to select the $right$ block for stitching. In addition, we design a on-the-fly search space optimizer to continuously reduce the size of the candidate block pool. Moreover, to reduce the energy consumption during block selection in local devices, we propose an energy coordinator to minimal the energy cost while meeting the deadline.  
% The experiment results show that
% FedStitch improves the model performance up to 20.93\%,
% effectively reduces the memory overhead up to 79.5\%,
% accelerates the network generation speed by up to 8.12$\times$,
% and achieves energy-saving up to 89.41\%.

In this paper, we introduce FedStitch, a novel FL approach tackling memory limitations and statistical diversity. FedStitch leverages pre-trained networks, breaking them into blocks for reassembly into new networks for specific tasks, eliminating the need for training and minimal data usage. This approach conserves computational resources and memory during backpropagation. To address performance drops from statistical diversity in FL, we implement a reinforcement learning-based algorithm for weighted block aggregation and introduce a real-time optimizer to narrow down block choices. Additionally, an energy coordinator reduces energy use during block selection, ensuring efficiency.  The experimental results demonstrate that FedStitch effectively avoids the memory requirements for training, improves the generated model accuracy, and optimizes energy consumption.

\section*{Acknowledgment}
This paper is supported by the Science and Technology Development Fund of Macau SAR (File no. 0081/2022/A2, 0123/2022/AFJ), MYRG-GRG2023-00211-10TSC-UMDF, SRG2022-00010-10TSC. Please ask  Dr. Li Li (llili@um.edu.mo) for correspondence.
% This paper is supported the Science and Technology Development Fund of Macau SAR 
% (File ), MYRG-GRG2023-00211-10TSC-UMDF, SRG2022-00010-10TSC. Please ask  Dr. Li Li (llili@um.edu.mo) for correspondence.

\vspace{12pt}

% \color{red}
% IEEE conference templates contain guidance text for composing and formatting conference papers. Please ensure that all template text is removed from your conference paper prior to submission to the conference. Failure to remove the template text from your paper may result in your paper not being published.

\end{document}